\documentclass[twoside,11pt]{article}

\usepackage{jmlr2e}
\usepackage{hyperref}

\usepackage{amsmath}

\usepackage{algorithm}
\usepackage{algpseudocode}

\usepackage{color}
\definecolor{vermilion}{rgb}{1,0.3,0}
\definecolor{venetianred}{rgb}{0.78,0.03,0.08}
\definecolor{tyrianpurple}{rgb}{0.4,0.01,0.24}

\newcommand{\Algref}[1]{Algorithm~\ref{#1}}
\newcommand{\aref}[1]{Appendix~\ref{#1}}
\newcommand{\eref}[1]{(\ref{#1})}
\newcommand{\fref}[1]{Figure~\ref{#1}}
\newcommand{\frefs}[1]{Figures~\ref{#1}}

\newcommand{\Fref}[1]{Figure~\ref{#1}}

\newcommand{\edity}[1]{{#1}}
\newcommand{\corry}[1]{{#1}}
\newcommand{\blindey}[1]{{#1}}

\newcommand{\setVar}{\boldsymbol{X}}
\newcommand{\bnet}{\mathcal{B}}
\newcommand{\graph}{\mathcal{G}}
\newcommand{\party}{\lambda}
\newcommand{\Party}{\Lambda}
\newcommand{\permy}{\pi}

\newcommand{\chainStep}{j}
\newcommand{\pars}{\theta}
\newcommand{\Pa}{\mbox{{\bf Pa}}}
\newcommand{\obs}{D}
\newcommand{\score}{S}
\newcommand{\order}{\prec}
\newcommand{\neibr}[1]{\#\mathrm{nbd}(#1)}
\newcommand{\neib}[1]{\mathrm{nbd}(#1)}
\newcommand{\Sord}{R}

\ShortHeadings{Partition MCMC for Inference on Acyclic Digraphs}{\blindey{Kuipers and Moffa}}
\firstpageno{1}

\begin{document}

\title{Partition MCMC for Inference on Acyclic Digraphs\blindey{\thanks{{R} code is available at \url{https://github.com/annlia/partitionMCMC}}}}

\blindey{\author{\name Jack Kuipers \email jack.kuipers@bsse.ethz.ch 
\\ \addr D-BSSE, ETH Zurich \\ Mattenstrasse 26 \\ 4058 Basel, Switzerland
\AND
\name Giusi Moffa \\ \addr Division of Psychiatry \\ University College London \\ London, UK}
}


\editor{\hspace{-1.5cm}{\color{white}\rule[-0.2cm]{2cm}{0.6cm}} \vspace{-1.8cm}}

\maketitle

\begin{abstract}
Acyclic digraphs are the underlying representation of Bayesian networks, a widely used class of probabilistic graphical models. Learning the underlying graph from data is a way of gaining insights about the structural properties of a domain. Structure learning forms one of the inference challenges of statistical graphical models.

MCMC methods\corry{, notably structure MCMC,} to sample graphs from the posterior distribution given the data are probably the only viable option for Bayesian model averaging. Score modularity and restrictions on the number of parents of each node \corry{allow} the graphs to be grouped into larger collections, which can be scored as a whole to improve the chain's convergence.  Current examples of algorithms taking advantage of grouping are the biased order MCMC, which acts on the alternative space of permuted triangular matrices\corry{,} and non ergodic edge reversal moves.

Here we propose a novel algorithm, which employs the underlying combinatorial structure of DAGs to define a new grouping. As a result convergence is improved compared to structure MCMC, while still retaining the property of producing an unbiased sample. Finally the method can be combined with edge reversal moves to improve the sampler further.
\end{abstract}


\section{Introduction}\label{intro}

A key question in applied statistics is to learn the relationship between a number of interacting entities and their surrounding environment.  \corry{Examples include} the many genes of a genome, the molecules in a cell, the different cells of an organism, demographics, economic and political factors in the dynamics of society, behavioural habits and biology in the development and progression of diseases. 
Most often\corry{,} not only \corry{is} the strength of the relationship between the variables \corry{unknown} but \corry{so too is} the connectivity structure itself\corry{,} and of great scientific interest. Think for \corry{instance} of gene regulatory networks \citep{friedmanetal00, husmeier03, friedman04, rauetal12} and cellular signalling pathways \citep{sachsetal05, ms08, hilletal12}, including their development over time \citep{husmeier03, rauetal12, hilletal12}.

Probabilistic graphical models provide a framework \corry{for characterising} the joint probability distribution of the variables in a domain and \corry{making} inference about features of interest. Bayesian networks are a popular class of probabilistic graphical models with directed acyclic graphs (DAGs) as \corry{their} underlying structure. The representation is such that the joint probability distributions of the nodes of the network can be written as a product of the conditional distribution of each node given its parents in the graph. In general no structural information is available and the learning process involves both estimating the graphical structure and the parameters which characterise the conditional probability distributions of each node on its parents, given a particular network topology.

A rather comprehensive overview of approaches for Bayesian network learning is given by \cite{art:DalySA2011}.  Structure learning is known to be a hard problem, especially due to the super-exponential growth of the DAG space when increasing the number of nodes. Broadly speaking the literature about structure learning can be divided \corry{into} two classes:  constraint-based methods, and score and search algorithms \citep[as discussed \corry{for example} in][]{bk:kf09}. 

The edges of a Bayesian network encode conditional independence relationships, which constitute the main ingredient of constraint based learning methods, such as the widely used PC algorithm  \citep{bk:sgs00, art:KalischB2007} \corry{which was} also recently implemented in the {\tt pcalg} {R} package \citep{art:KalischMCMB2012}.  \edity{Starting from a complete skeleton, decisions about whether edges should be deleted are made recursively based on tests of conditional independence.  The tests start from pairwise comparisons and increase in complexity.} By their very nature\corry{,} methods building on this strategy are sensitive to local errors of the tests and to the order in which they are run \citep{cm14}, but they tend to scale relatively well with the dimension. 

On the other side of the spectrum are score and search algorithms, which rely on the definition of a measure of fit of a graphical model to the observed data. The fit is typically evaluated by scoring the entire network, with the drawback that the method requires exploration of the large network space. 

In an attempt to exploit the strengths of each approach, hybrid solutions have also been proposed, as for example the max-min-hill-climbing \corry{method} of \cite{tba06}\corry{. Their algorithm} first learns a skeleton of admissible edges in the network and then proceeds by a greedy hill-climbing restricted to the space of structures compatible with the learned skeleton. Hybrid methods take advantage of the constraint based ideas to perform -- possibly rather conservative -- significance tests in a first phase\corry{,} only to reduce the search space of a second \corry{score and search} phase.

The direction of the edges in a Bayesian network are interesting from a causal perspective\corry{,} since the connections learned from observational data may help unveil unknown causal relations. Due to the great potential for shedding light on important scientific questions, the possibility of deriving causal statements from DAGs has \corry{had great appeal} ever since the causal interpretation of Bayesian networks was proposed \citep{pv91,pearl00}. Caution must however be taken in interpreting DAGs causally \citep{dawid10}, since it is only justified under strict and \corry{quite} likely untestable assumptions, such as not having unmeasured confounders. Moreover\corry{,} graphical structures encoding the same conditional independencies cannot be distinguished from observational data, so that \corry{even} in the best case scenario only the equivalence class of all the networks describing the same probability distribution can be inferred. 

Recently \cite{mkb09} proposed an interesting extension of the intervention calculus of \cite{pearl00} to scenarios where the underlying DAG is unknown and only an equivalence class can be learned from data, \corry{while} retaining the strict assumptions of faithfulness and absence of unobserved confounders. A combination of observational and interventional data \corry{may however} help to distinguish between the models of an equivalence class (\citealp{cy99,em07,hb13}; see also \citealp{friedmanetal00} for a concise discussion about the discovery of causal patterns from observational data). 

The focus of our work is on score and search methods, and in particular on MCMC methods for the graph space exploration. The main advantage of MCMC approaches with respect to methods based on greedy searches and other optimisation algorithms is that they can provide a collection of samples from the posterior distribution of the graph given the data. This means that inference can be made in the spirit of Bayesian model averaging, since the expectation of given network features, such as the posterior probability of an individual edge, can be estimated by averaging over the \edity{sample} \citep{my95}.  The possibility \corry{of conducting} Bayesian model averaging is especially important in high dimensional domains with sparse data, where no single best model can be clearly identified, so that \corry{relying} on the best scoring model to \corry{perform} inference \corry{is unjustified}.

The first MCMC algorithm over graph structures is due to \cite{my95}, later refined by \cite{gc03}. To improve on the mixing and convergence, \cite{fk03} suggested to build a Markov chain on the space of node orders instead, at the price of introducing a bias in the sampling.  For smaller systems, space and time complexity are such that an efficient option is given by dynamic programming \citep{ks04}, which can \corry{also} be used to extend the proposals of standard structure MCMC in a hybrid method \citep{em07}.  Within the MCMC approach, to avoid the bias \corry{in order MCMC}, while keeping reasonable convergence, \cite{gh08} more recently proposed a new edge reversal move combining ideas both of standard structure and order MCMC.

In this paper we present a novel MCMC algorithm designed on the combinatorial structure of DAGs, with the advantage of improving convergence with respect to structure MCMC\corry{.  At the same time it still provides} an unbiased sample since it acts directly on the space of DAGs. \corry{It} can also be combined with the algorithm of \cite{gh08}, in place of structure, to promote convergence.

Bayesian networks provide a valuable tool \corry{for gaining} insights from observational data about the mechanism underlying the data generating process\corry{,} and help to design experiments which can validate observational findings.  \corry{Efficient} tools for structure learning \corry{are therefore} highly important. While there are a number of software packages, such as {\tt pcalg} \citep{art:KalischMCMB2012} and {\tt bnlearn} \citep{scutari10} in {R}, for constraint based learning methods, \corry{publicly} available tools for MCMC based structure learning algorithms are less common, with the \corry{possible} exception of {\tt BDAGL} \citep{BDAGL07}.   \corry{In tandem} with dynamic programming approaches, both standard structure, and MCMC in the space of orders are implemented in {\tt BDAGL}, but not the more recent new edge reversal move of \cite{gh08} and the package is written for \edity{{Matlab}}.

With this \corry{paper} we provide {R} code for structure \citep{my95} and order \citep{fk03} MCMC, \corry{together with} the algorithm of \cite{gh08} and our newly proposed partition MCMC.

\section{Terminology and notation of Bayesian networks}

Given a set of $n$ random variables $\setVar = \{ X_1, \ldots, X_n\}$ we are interested in characterising their joint probability distribution by means of a directed graphical model. 
A Bayesian network $\bnet = (\graph, \pars)$ can be fully specified by associating a set of parameters $\pars$ to a directed acyclic graph $\graph$ whose nodes are the random variables in $\setVar$. The parameters $\pars$ specify a conditional probability distribution $P(X_i \vert \Pa_i)$ for each variable $X_i$ given the set of its parents $\Pa_i$ in the graph $\graph$. From the \emph{Markov assumption} \citep{bk:kf09} that each variable $X_i$ is independent of its non-descendants given its parents in the graph $\graph$, it follows that the joint probability distribution described by the Bayesian network $\bnet$ factorizes as
\begin{equation} \nonumber
P( X_1, \ldots, X_n ) = \prod_i^n P(X_i \vert \Pa_i) \, .
\end{equation}
The random variables in $\setVar$ can be discrete, continuous\corry{,} or a mixture of both. \corry{In} our examples we will focus on continuous variables with a multivariate Gaussian distribution (\citealp{gh02}; see also the correction in \citealp{cr12,kmh14}).

In practice we may want to learn a Bayesian network $\bnet$ which best explains a set of independent observations $\obs$ from the distribution of the variables in $\setVar$. The estimation of a Bayesian network breaks up into two steps \citep{bk:cowelletal07}\corry{,} referred to as structure and parameter learning. Given the data $\obs$ we wish to learn the structural dependence of the variables in $\setVar$ encoded by a DAG $\graph$ and estimate a set of parameters $\pars$ for the corresponding $\graph$. The focus of our work here is on the structure learning, in the context of search and score algorithms. 

It is well known\corry{,} however\corry{,} that the structure of the network, or in other words the DAG, is identifiable \corry{only} up to an equivalence class. All DAGs in an equivalence class encode the same probability distribution. Equivalent DAGs share the same skeleton, or undirected underlying graph, and the same v-structures \citep[two parents with the same child and no direct edge between them,][]{vp90}. 
In practice, even from perfect \edity{or noiseless} data, when all conditional independencies are known exactly, we can only learn the common features of an equivalence class \citep{mkb09}, which can be represented by a completed partially directed acyclic graph (CPDAG) or an essential graph \citep{amp97}. 

The scoring functions for the graphical structure are typically derived from a Bayesian approach, where the score for a DAG $\graph$ is defined as its posterior probability given the data $\obs$
\begin{equation} \nonumber
P( \graph \vert \obs) \propto P(D \vert \graph) P(\graph) \, ,
\end{equation}
with $P(\graph)$ a prior distribution over graphical structures. The marginal likelihood $P(D \vert \graph)$ is obtained by integrating the likelihood function $P(D \vert \graph, \theta)$ over the parameter prior $P(\theta \vert \graph)$ for a given graph $\graph$ and over the parameter space $\Theta$. When the prior satisfies the conditions of structure modularity, parameter independence and parameter modularity as defined by \cite{hg95}, and if the data is complete\corry{,} the score is structure equivalent and decomposable \citep{friedmanetal00,fk03}
\begin{equation} \nonumber
P( \graph \vert \obs) \propto P(D \vert \graph) P(\graph) = \prod_i \score(X_i, \Pa_i \vert D) \, ,
\end{equation}
where $\score$ is a score function depending only on the node variable $X_i$ and its parents. 
The decomposability is important from an implementation point of view, since it means that during a structure search, only the score of the nodes whose parents change with respect to the previously scored structure needs to be reevaluated. Since the number of possible DAGs grows super-exponentially with the number of vertices\corry{,} an exhaustive search \corry{quickly} becomes impracticable even for \corry{a} moderate number of nodes. State of the art approaches rely on approximate solutions, \corry{in} particular from simulation methods based on MCMC.

\section{Structure learning of Bayesian networks by MCMC methods}

Before presenting our algorithm\corry{,} we briefly review the current state of the art of MCMC methods for structure learning of Bayesian networks. The most common strategies rely either on elementary moves in the graph structure, involving a single edge, or on sampling in the space of node orders, but leading to biased samples.
The novelty of our approach consists in starting from the combinatorial representation of DAGs to build an efficient MCMC scheme directly on the space of DAGs. 

\subsection{Structure MCMC}

The classical MCMC method for learning the underlying structure of Bayesian networks dates back to \cite{my95} and is referred to by \cite{fk03} as structure MCMC. It is based on the simple idea of building a Markov chain on the space of graphical models, such that its stationary distribution is the posterior distribution $P( \graph \vert \obs)$ of the network $\graph$ given the data $\obs$. The simplest procedure constructs a random walk by means of the simple operations of \emph{addition and deletion} of single edges in a Metropolis Hastings algorithm. 

Given a DAG $\graph_\chainStep$ at iteration $\chainStep$, find the neighbourhood $\neib{\graph_\chainStep}$ of all DAGs with one edge added or deleted (and including $\graph_\chainStep$ itself).  Sample a new graph $\graph'$ uniformly from this neighbourhood with proposal probability
\begin{equation} \nonumber
	q(\graph' \vert \graph) = 
		\left \{
			\begin{array}{cc}
				\frac{1}{\neibr{\graph_\chainStep}} & \mbox{if } \graph' \in \neib{\graph_\chainStep} \\
				0 & \mbox{otherwise}
			\end{array}
		\right. \, ,
\end{equation} 
as defined by \cite{my95}.

The acceptance probability for the proposed graph in a Metropolis Hastings algorithm is then
\begin{equation} \nonumber
\rho =  \min \left \{ 1 , \frac{q(\graph_\chainStep \vert \graph') P( \graph' \vert \obs) }{q(\graph'\vert\graph_\chainStep) P( \graph_\chainStep \vert \obs)} 
		\right \}
	 = \min \left \{ 1 , \frac{\neibr{\graph_\chainStep} P( \graph' \vert \obs) }{\neibr{\graph'} P( \graph_\chainStep \vert \obs)} 
		\right \} \, ,
\end{equation}		
so that the next state in the chain $\graph_{\chainStep+1} = \graph'$ with probability $\rho$ and $\graph_{\chainStep+1} = \graph_\chainStep$ otherwise.

Once the chain converges\corry{,} it provides a sample $\graph^\star$ of a graphical structure from the posterior distribution $P( \graph \vert \obs)$, providing the means to conduct inference based on Bayesian model averaging.  Typically this means collecting a (correlated) sequence of DAGs from one or several MCMC chains\corry{,} after a \emph{burn-in} period to decouple from the starting point of the chain. Since processing the DAGs after they have been sampled \corry{can be} quite computationally expensive, \emph{thinning} is often appropriate \citep{le12}.

By modifying the original algorithm of \cite{my95} \corry{to} include the possibility of \emph{reversing} an edge\corry{,} the convergence speed can be greatly improved \citep{gc03}.  Moreover, rejection sampling from a larger fixed sized neighbourhood can be employed to avoid explicitly calculating the neighbourhood\corry{,} which can further speed up the implementation \citep{gc03}.  Even with these improvements, structure MCMC can still struggle to converge for relatively small DAGs. \edity{In \aref{BHexample} for example, severe convergence difficulties are already apparent for DAGs with 14 nodes.}

To assess the accuracy of structure learning algorithms, \cite{tba06} introduced the structural Hamming distance (SHD) on CPDAGs. The SHD corresponds to the number of simple operations, namely additions and deletions of both directed and undirected edges, or reversing of directed edges, required to go from one CPDAG to the other.

Although the SHD is calculated on CPDAGs rather than DAGs (to avoid penalising statistically non identifiable differences), one could say that the SHD is the rational behind structure MCMC.  Namely, proposals are limited to graphical structures with a SHD of 1 from the current graph. Interestingly\corry{,} what complicates things for exploring the space of graphical structures is not only the super-exponential growth of their space size, but also the fact that the behaviour of the SHD does not necessarily correspond to similar behaviours in the likelihood landscape. 
The score of a network \corry{may} in fact change substantially when performing small changes to the structure \citep{fk03}, or it may\corry{,} on the other hand\corry{,} vary very little when making more important modifications to the network structure. 

Intuition suggests that it should be possible to define more efficient MCMC schemes by making proposals which more closely reflect the likelihood landscape. Defining suitable distances between graphical structures to help \corry{design} better schemes is unfortunately not straightforward.  In the context of causal graphs a distance was recently defined \citep{pb13}, though purely based on a graphical criterion, directly related to the network interpretation in terms of differences in the causal inference statements deriving from each structure.

\subsection{Order MCMC}

To overcome the slow mixing property of structure MCMC, \citeauthor{fk03} introduced the order MCMC algorithm \citep{fk03}.  Key is the concept of \corry{a} node ordering $\order$ which is simply a permutation of the $n$ node labels. The nodes are essentially lined up in a chain and labelled according to the given permutation. The DAGs belonging to the corresponding order are such that each node may only have parents from further up the chain, or following it in the ordering.

For example if for $n=3$ we choose the ordering $3,1,2$ then node $3$ could have either of the others (or both or none) as parents.  Node $1$ may only have node $2$ (or none) as parents while node $2$ is forced to have no parents.  The possible choices of parents can be represented as an adjacency matrix where the rows and columns are labelled according to the order
\begin{equation}\nonumber
\begin{array}{c|ccc}  & \;\;3\; & \;1\; & \;2\; \\
\hline
\;3\; & 0 & 0 & 0 \\
\;1\; & \{0,1\} & 0 & 0\\
\;2\; & \{0,1\} & \{0,1\} & 0 
            \end{array} 
\end{equation}
so that only the lower triangular elements can differ from 0.  Each choice for the lower triangular elements is a different DAG and there are therefore 8 different DAGs compatible with this ordering of $3$ nodes.

The great insight of \cite{fk03} \corry{was} that if the score is modular, and we precompute the score of each node's possible parent sets\corry{,} we can efficiently sum the scores of all the DAGs compatible with a particular node ordering.  For each node we simply sum the scores of all the parent sets that do not include nodes preceding it in the ordering. The product of the node score sums over possible parent sets provides a score $\Sord(\order \vert \obs)$ of the entire order $\order$ given the data $\obs$
\begin{equation}\label{orderScore}
\Sord(\order \vert \obs) = \sum_{\graph \in \order} P( \graph \vert \obs) \propto \prod_{i=1}^{n} \sum_{\Pa_i \in \order} \score(X_i, \Pa_i \vert D) \, .
\end{equation}
However, for each node the number of parents sets is $2^{n-1}$ which all need to be scored and searched through to find the score of each order.  To prevent the exponential complexity as $n$ increases, a hard limit $K$ on the size of the parent sets is typically introduced to reduce the complexity of scoring each node to order $n^K$. \edity{A low threshold could exclude highly scoring DAGs while increasing $K$ too highly could increase the computational cost without a concomitant improvement in the DAGs included.  One therefore aims to select the lowest value which still encapsulates the bulk of the posterior weight.}

\corry{A} Markov chain \corry{can then be constructed} on the smaller space of node orders rather than the space of all DAG structures. A chain with stationary distribution proportional to $\Sord(\order \vert \obs)$ can be produced by a Metropolis Hastings algorithm with acceptance probability
\begin{equation} \label{orderMHaccprob}
\rho =  \min \left \{ 1 , \frac{ q(\order_\chainStep \vert \order') \Sord( \order' \vert \obs) }{ q(\order' \vert \order_\chainStep) \Sord( \order_\chainStep \vert \obs)} \right \} \, ,
\end{equation}
where $q(\order' \vert \order)$ is the probability of proposing a move to $\order'$ from $\order$, and can be any move in the space of permutations or orders \citep[see][for some examples]{fk03}.  The simplest move consists in flipping two nodes in the order while leaving the other unchanged.  This is symmetric so that the $q$ terms in \eref{orderMHaccprob} cancel.

Upon convergence\corry{,} order MCMC provides a sample of an order $\order^{\star}$ from a distribution proportional to the score $\Sord(\order \vert \obs)$ over the space of $n!$ possible orders of the nodes of the graphical structure. Given a sampled order, one can sample a DAG by sampling the parents of each node independently according to the scores of its permissible parent sets.

By grouping together and averaging the score over so many DAGs\corry{,} the convergence properties of the MCMC chain on the much smaller space of orders are vastly improved compared to structure MCMC \citep{my95,gc03}.

The convergence improvement\corry{,} however\corry{,} only works by ignoring the combinatorial structure of DAGs and working on the much simpler space of permuted triangular matrices.  With $n$ nodes there are $2^L$ DAGs consistent with each order with $L=n(n-1)/2$ the number of lower triangular elements, since each can take one of two values.  There are also $n!$ orders giving a total of $n!2^L$ permuted lower triangular matrices.
The number of DAGs, $a_n$, \corry{is} exponentially smaller, as can be seen from the asymptotic behaviour \citep{robinson70,robinson73,stanley73}
\begin{equation} \nonumber
a_n\sim \frac{n!2^{L}}{Mq^n} \, ,
\end{equation}
where $M=0.574\ldots$ and $q=1.48\ldots$ are constants.  The number of orders each DAG belongs  to is therefore exponentially large on average (it can range from 1 to $n!$).  The fact that a given graph $\graph$ may be consistent with more than one order induces a bias in the posterior distribution defined on the space of graphical structures.

This bias arises since the expression in equation \eref{orderScore} does not exactly correspond to the posterior distribution $P(\order \vert \obs)$ which can be written instead as
\begin{equation} \nonumber
P(\order \vert \obs) 
	= \sum_{G} P(\graph, \order \vert \obs) 
	= \sum_{G \in \order} P(\order \vert \graph) P(\graph \vert \obs) \, .
\end{equation}
The difference with respect to \eref{orderScore} consists of the term $P(\order \vert \graph)$, which is the inverse of the number of orders the DAG $\graph$ is consistent with.  Neglecting this term in the order MCMC algorithm then weights DAGs by the number of orders they belong to, resulting in the aforementioned bias.

This weighting can also be seen as a consequence of placing a prior on orders $P(\order)$.  The prior on graphs is then determined as $P(\graph) = \sum_\order P(\graph \vert \order)P(\order)$ so each DAG obtains a contribution from each order it can belong to.  Attempts to remove the bias via importance sampling \citep{ew08} can help for small graphs, but they struggle as the size increases due to the exponential number of orders DAGs can be consistent with on average.

\subsection{New edge reversal moves}

Since the bias is the main problem with order MCMC, while the slow convergence is the main limit of structure MCMC, \cite{gh08} introduced a novel edge reversal move into structure MCMC in an attempt to overcome both difficulties. Their new move also relies on the key feature of order MCMC\corry{,} of combining the score of many possible parent sets. Moreover, when an edge is reversed, the parents of the two nodes that it connects are resampled according to their score. Since the jumps are chosen according to their score, the chain moves to more probable DAGs more quickly, which vastly improves the convergence of structure MCMC. The edge reversal move typically results in the proposal of DAGs with a SHD bigger than 1 from the current graph in the chain, so that larger jumps than those of structure MCMC are possible.

Since the edge reversal operation requires the knowledge of the scores of many possible parent sets, again a hard limit $K$ on the size of such sets is usually introduced as in order MCMC. Moreover by itself the new edge reversal move is not ergodic in the space of DAGs. To overcome \corry{this} problem\corry{,} the complete sampling algorithm of \cite{gh08} combines the new move with an underlying structure MCMC chain in a mixture, where a new reversal move is proposed with a given probability $p_{\rm rev}$ and a classical structure move with probability $1-p_{\rm rev}$. Since both structure MCMC and the new edge reversal move are unbiased, the algorithm based on their mixture is also unbiased in the space of DAGs. 

At the same time as avoiding the bias inherent in order MCMC the algorithm of \cite{gh08} exhibits much better convergence than structure MCMC. One may try to combine all the currently available methods to find the most efficient way of sampling DAGs according to their posterior probabilities.  A valid strategy consists of using order MCMC to find a graph from which to start a chain, which is then continued with the mixture of structure MCMC and edge reversal.

\section{Partition MCMC}

In \corry{the current paper} we propose a novel MCMC method which \corry{adheres to} the philosophy of order MCMC but which respects the combinatorial structure of DAGs.  In particular\corry{,} we define a MCMC algorithm on the space of node partitions, essentially a subdivision of orders necessary to avoid over representing certain DAGs.  The subdivision will in general slow the convergence compared to order MCMC, but by acting directly on the space of acyclic digraphs our algorithm does not suffer from bias.  \corry{As} for edge reversal, one may run order MCMC to start the chain and then continue with partition MCMC to remove the bias.

\corry{The} grouping of DAGs into partitions means that the convergence is much improved with respect to structure MCMC.  Moreover\corry{,} we can efficiently combine our method with the new edge reversal move of \cite{gh08} and improve upon their MCMC sampler.

\subsection{Outpoints}

\begin{figure}
  \centering
  \includegraphics[width=0.65\textwidth]{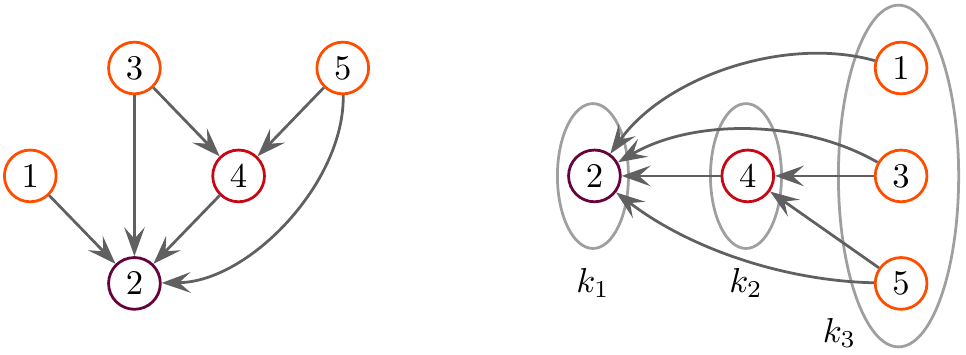}
  \caption{By collecting the outpoints at each step into a set, the DAG on the left can be redrawn according to the partition $[1,1,3]$ where now edges are only allowed to come from the right.}
  \label{dagexample}
\end{figure}

DAGs, since they do not admit cycles, must have at least one \emph{outpoint}, defined as a node with no incoming arcs. \edity{Outpoints are also known as \emph{sources}.} In \fref{dagexample} for example nodes $1,3$ and $5$ are outpoints.  If these, and their outgoing edges, are removed from the graph then we are left with a smaller DAG.  A single outpoint is then left as node $4$, which when removed leaves node $2$ as the \corry{sole} outpoint.  This property allows DAGs to be built recursively and enumerated \citep{robinson70,robinson73}, which also means they can be sampled uniformly \citep{km13}.

If we combine the outpoints at every stage into $m$ sets each of size $k_i$ then since all nodes are placed somewhere\corry{,} $\sum_{i=1}^{m} k_i = n$ and we have partitioned $n$ into $[k_1,k_2,...k_m]$.  For \corry{example} in \fref{dagexample}\corry{,} we have the following three sets: $\{2\}$, $\{4\}$ and $\{1,3,5\}$ and the partition $[1,1,3]$.  \blindey{The partition order is reversed from \cite{km13}.}  

When arranging the partitioned sets of nodes in groups from left to right, edges are only allowed to come from sets, or partition elements, further to the right.  The arrangement is similar to the ordering used in order MCMC, but there are two additional restrictions:
\begin{itemize}
\item nodes in the same partition element are not allowed to be connected to each other (otherwise they would not be \corry{concurrent} outpoints) 
\item nodes must be connected to by at least one directed edge from the nodes in the \edity{adjacent} partition element to the right (otherwise they would be outpoints at an earlier stage).
\end{itemize}
For example, in \fref{dagexample} node $2$ must receive an edge from node $4$. 

\subsection{MCMC for uniform sampling}

The number of DAGs belonging to a given partition follows from the number of edge possibilities.  Let $S_j=\sum_{i=j}^{m}k_i$ then
\begin{equation} \nonumber 
a_{[k_1,\ldots,k_m]} = \frac{n!}{k_1! \ldots k_m!} \prod_{j=1}^{m-1} \left(2^{k_{j+1}}-1\right)^{k_{j}}\prod_{j=1}^{m-2} 2^{k_{j}S_{j+2}} \, .
\end{equation}
The first combinatorial term is the number of ways of distributing the $n$ nodes into the $m$ partition elements of size $k_1,\ldots , k_m$.  Basically, permuting the nodes labels inside a partition element has no effect on the set of DAGs consistent with the partition.  The second term counts the number of ways in which the nodes in each partition element can receive edges from \edity{the adjacent element to the right}, where subtracting 1 inside the bracket excludes the case when the nodes receive no edges.  The final term is the number of possible edges from nodes in partition elements even further right.  \edity{The number of DAGs in a partition therefore varies from 1 for the partition with a single element up to $n!2^{L-n+1}$ for the partition with $n$ elements.}

By assigning each partition $P$ a score $a_P$, a MCMC scheme was previously introduced in the space of partitions to sample DAGs uniformly \citep{km13}.  In particular the moves were chosen so that the ratio of scores $\frac{a_{P'}}{a_P}$ in the acceptance probability simplified and was as cheap as possible to evaluate.  Here we extend the method to sample from a posterior distribution in the context of Bayesian inference for directed graphical models.  Namely, a MCMC sampling procedure can be built from the combinatorial structure of DAGs, which \corry{comprises}:
\begin{itemize}
 \item an ordered partition of $n$, namely $\party=[k_1,k_2,...k_m]$ with $\sum k_i = n$ 
 \item a permutation $\permy$ on the node labels
 \item edges connecting the nodes (with certain restrictions).
\end{itemize}

\subsection{Scoring partitions}

For a given partition and permutation, permuting the labels inside a partition element does \corry{not matter,} so \corry{an} ordering can be fixed.  Denote by $\permy_\party$ a single representative of the equivalent permutations with respect to the partition $\party$ so that the pair $(\party,\permy_\party)=\Party$ can be viewed as a labelled partition.  

Analogously to order MCMC, for a given labelled partition $\Party$ and with a modular score we can score all permissible parent sets for each node.  From the list of all parent sets we exclude any with a member in the same partition element or in one further left.  Only parent sets with at least one member in the partition element immediately to the right need to be included.  For example, if $\party=[1,2,2]$ and $\permy_{\party} = 2,3,4,1,5$ as in \fref{basicmove}(c), we can look at the possible parent sets for node $3$.  Nodes $2$ and $4$ are excluded as possible parents while at least one of $1$ and $5$ must be included, resulting in three possible parent sets.

By summing the scores of the permissible parent sets for each node (and multiplying the sums for all nodes), we treat all the possible edge combinations and hence combine the score of all DAGs consistent with the given labelled partition $\Party$.  The total score so obtained coincides with the posterior probability of the labelled partition
\begin{equation} \label{partitionscoreeq}
P(\Party \vert \obs) 
	= \sum_{\graph} P(\Party \vert \graph, \obs)  P( \graph \vert \obs) 
	\equiv \sum_{\graph \in \Party} P( \graph \vert \obs) 
	\propto \prod_{i=1}^{n} \sum_{\Pa_i \in \Party} \score(X_i, \Pa_i \vert D) \, ,
\end{equation}
and a MCMC chain can be built on the joint space of partitions and permutations.  Each MCMC move needs to propose a new labelled partition $\Party'$ which is accepted with probability
\begin{equation} \label{partitionacceptratio}
\rho =  \min \left \{ 1 , \frac{\neibr{\Party} P( \Party' \vert \obs) }{\neibr{\Party'} P( \Party \vert \obs)} 
		\right \} \, ,
\end{equation}		
where the neighbourhood needs to be calculated for each move type.

\subsection{Basic move}

\begin{figure}
  \centering
  \includegraphics[width=\textwidth]{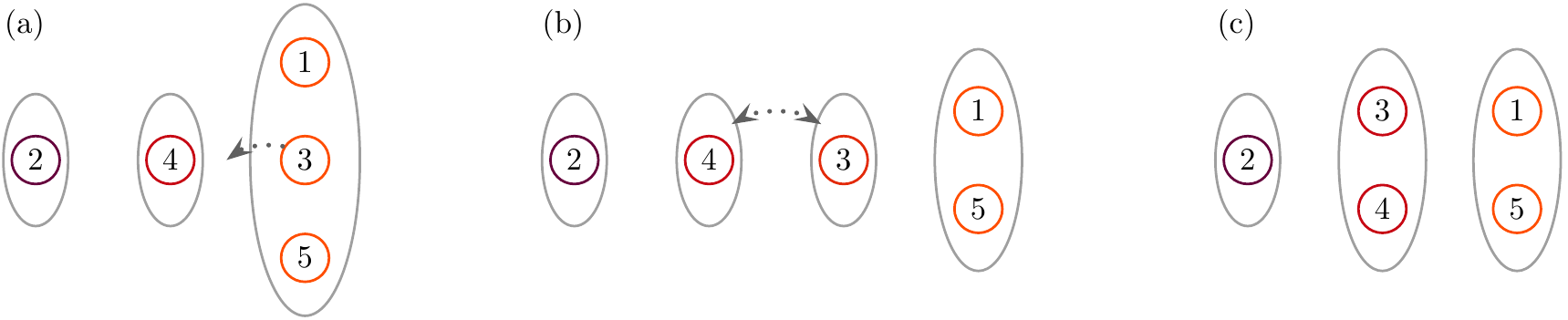}
  \caption{From the partition $\party=[1,1,3]$ in \fref{dagexample} drawn in (a) splitting the partition element containing nodes $\{1,3,5\}$ into two by moving node 3 into a new partition element on the left leads to the labelled partition in (b) with $\party=[1,1,1,2]$.  Further combining the two partition elements containing nodes 4 and 3 leads to the partition in (c) with $\party=[1,2,2]$.}
  \label{basicmove}
\end{figure}

The basic move consists of splitting a partition into two or joining two \edity{adjacent} ones.  \blindey{In \cite{km13} this move was performed by a mapping to binary sequences.}  As an example, in \fref{dagexample} which is redrawn in \fref{basicmove}(a), node $4$ may be joined to node $2$ or the set $\{1,3,5\}$ and the partition element containing $\{1,3,5\}$ may be split in two.  If we do split $\{1,3,5\}$ into two by separating off node $3$ to the left, we move from \fref{basicmove}(a) to \fref{basicmove}(b).  Further joining nodes $3$ and $4$ gives the labelled partition in \fref{basicmove}(c).    

However when splitting an element of size $k$ into two elements of size $c$ and $k-c$ respectively there are $\binom{k}{c}$ ways to separate the nodes and correspondingly update the permutation.  With $(m-1)$ ways of joining partition elements, the size of the neighbourhood is then
\begin{equation} \nonumber
m-1 + \sum_{i=1}^{m} \sum_{c=1}^{k_i-1}\binom{k_i}{c} = m-1+\sum_{i=1}^{m} \left(2^{k_i}-2\right) = - m -1 + \sum_{i=1}^{m}2^{k_i} \, .
\end{equation}
\edity{We can therefore sample from the neighbourhood uniformly as summarised in \Algref{proposeLambda}. This then allows us to define the full MCMC scheme for the basic move in \Algref{basicMCMCmove}.}

\begin{algorithm}[t]
\caption{\edity{Sample a proposal partition $\Party'$ from the neighbourhood of $\Party$}}\label{proposeLambda}
\begin{algorithmic}
\State \textbf{input} An ordered partition $\Party$
\State Sample an integer $j$ uniformly from $1:\neibr{\Party}$ where
\begin{equation} \nonumber
\neibr{\Party}=m-1 + \sum_{i=1}^{m} \sum_{c=1}^{k_i-1}\binom{k_i}{c}
\end{equation}
with $m$ the number of partition elements in $\Party$ and $k_i$ the size of element $i$
\If{$j<m$}
\State Join partition elements $j$ and $j+1$ of $\Party$ to form $\Party'$
\Else
\State Find minimum $i^{\star}$ such that
\begin{equation} \nonumber
j \le m - 1+ \sum_{i=1}^{i^{\star}} \sum_{c=1}^{k_i-1}\binom{k_i}{c}
\end{equation}
\State Find minimum $c^{\star}$ such that
\begin{equation} \nonumber
j \le m - 1+ \sum_{i=1}^{i^{\star}-1} \sum_{c=1}^{k_i-1}\binom{k_i}{c} + \sum_{c=1}^{c^{\star}}\binom{k_{i^{\star}}}{c}
\end{equation}
\State Sample $c^{\star}$ nodes from partition element $i^{\star}$ in $\Party$
\State Split them off into a new partition element on the left to form $\Party'$
\EndIf
\State \textbf{return} $\Party'$
\end{algorithmic}
\end{algorithm}

\begin{algorithm}[t]
\caption{\edity{Basic MCMC in the partition space}}\label{basicMCMCmove}
\begin{algorithmic}
\State \textbf{input} Chain length $T$ and an initial ordered partition $\Party_0$
\For{$t=1$ to $T$}
\State Sample $\alpha$ uniformly from $(0,1)$
\If{$\alpha<0.01$} \Comment{Small probability to stay still}
\State $\Party_{t}=\Party_{t-1}$
\Else
\State Sample a proposal partition $\Party'$ from $\Party_{t-1}$ using \Algref{proposeLambda}
\State Sample $\alpha$ uniformly from $(0,1)$
\If{$\alpha<\frac{\neibr{\Party} P( \Party' \vert \obs) }{\neibr{\Party'} P( \Party \vert \obs)} $}
\State $\Party_{t}=\Party'$
\Else
\State $\Party_{t}=\Party_{t-1}$
\EndIf
\EndIf
\State Sample DAG $\graph_t$ from $\Party_t$ weighted following $P( \Party_{t} \vert \obs)=\sum_{\graph \in \Party_{t}} P( \graph \vert \obs)$
\EndFor
\State \textbf{return} $\{\graph_t\}$
\end{algorithmic}
\end{algorithm}

When joining two partition elements, only the nodes from the partition element on the left and its \edity{adjacent element} further left need to be rescored.  Similarly, when splitting a partition element, only the nodes in the newly formed element on the left, and its \edity{adjacent element} further left need to be rescored.  

\subsection{Sampling DAGs}

Splitting and joining partitions are inverse moves of each other, \edity{so the moves are reversible}. Since all partition elements can be joined into a single one in up to $(n-1)$ steps and then separated out into a new labelled partition in up to $(n-1)$ further steps, the chain is certainly irreducible after $2(n-1)$ steps and introducing a small probability of staying still\corry{,} will avoid the possibility of a periodic chain. \edity{Finally \eref{partitionacceptratio} ensures detailed balance since 
\begin{eqnarray}
P( \Party \vert \obs) P(\Party \to \Party') &=& \frac{P( \Party \vert \obs)}{\neibr{\Party}} \min \left \{ 1 , \frac{\neibr{\Party} P( \Party' \vert \obs) }{\neibr{\Party'} P( \Party \vert \obs)} 
		\right \}  \nonumber \\
		&=& \min \left \{  \frac{P( \Party \vert \obs)}{\neibr{\Party}} , \frac{P( \Party' \vert \obs) }{\neibr{\Party'} }  \right \} = P( \Party' \vert \obs) P(\Party' \to \Party) \, . \nonumber
\end{eqnarray}
The conditions to sample labelled partitions from the posterior $P( \Party \vert \obs)$ are therefore satisfied.}

\edity{Since $P( \Party \vert \obs)=\sum_{\graph \in \Party} P( \graph \vert \obs)$ as in \eref{partitionscoreeq}, from each sampled labelled partition $\Party$ we sample a single DAG, $\graph \in \Party$, weighted according to $P(\graph \vert \obs)$ to obtain DAGs sampled from their posterior. This DAG sampling step is then included in the MCMC scheme, for example with the basic partition move as outlined in \Algref{basicMCMCmove}.}

\subsection{Additional partition move}

To change the size of a partition element however, it is necessary \corry{first to} split nodes off and then join them to a different partition element.  For example, going from \fref{basicmove}(a) to \fref{basicmove}(c) takes two basic move steps.  A partition move could also be considered whereby we swap nodes directly from one partition element to another.

\begin{figure}
  \centering
  \includegraphics[width=\textwidth]{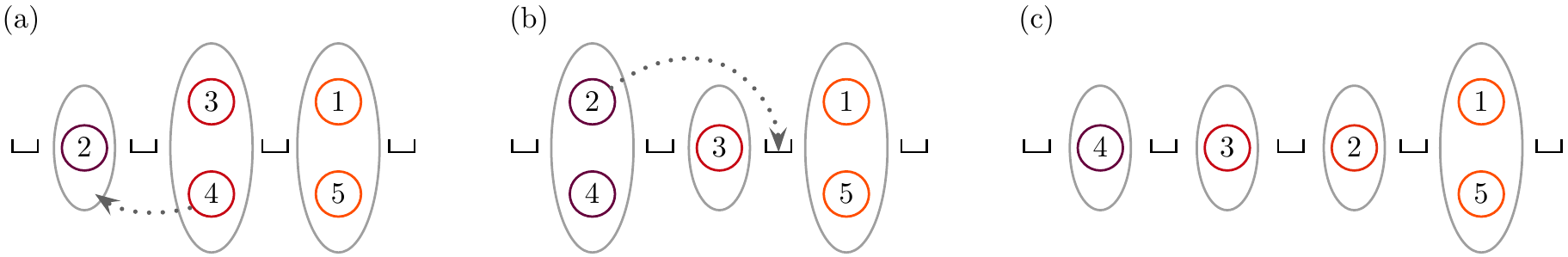}
  \caption{Redrawing the partition with $\party=[1,2,2]$ from \fref{basicmove}(c) with spaces either \corry{side} of the partition elements highlighted as in (a), we first sample node 4 and choose to move it into the partition element containing node 2 to arrive at the partition in (b) with $\party=[2,1,2]$.  Then we sample node 2 itself and choose to move it to a new partition element creating the partition in (c) with $\party=[1,1,1,2]$.}
  \label{partitionmove}
\end{figure}

\corry{It is possible to} move one node at a time.  For example, each node may move either to a different partition element or to one of the gaps \corry{in} between (or outside) to create a new partition element containing a single node.  In the partition of \fref{basicmove}(c) with $m=3$ elements, there are a total of four gaps as highlighted in \fref{partitionmove}(a).  To \corry{make a} move, \corry{first} sample a node uniformly from the $n$ available.  \corry{Then place that node in} one of the $(m+1)$ gaps or $(m-1)$ different partition elements\corry{, also chosen uniformly}.  In the example of \fref{partitionmove}\corry{,} we first sample node 4 to move to the first partition element, then node 2 to move to the third gap.  The neighbourhood of the move given the partition is simply $2mn$.

However, moving one node from a partition element containing two nodes into the gap to the right, or the other node into the gap to the left, leads to the same labelled partition.  Likewise moving a node from a partition element containing a single node into the gap either side, the partition remains unchanged.  To marginally improve the convergence, the latter possibility is excluded\corry{,} and nodes from partition elements \corry{with two nodes} are not allowed to move into the gap immediately to their left.  The neighbourhood calculation also needs to be adjusted to account for these exclusions.

As a node can be moved anywhere in the partition, a larger number of nodes might need to be rescored at each step. However, this move can be weighted relative to the basic move to keep the average number of nodes that need to be rescored down to around 4.  This is independent of $n$\corry{,} which keeps the complexity of the scheme lower.

\subsection{Permutation moves}

\corry{Currently, exploring} the space of permutations requires us to appropriately combine and split partition elements.  We could \corry{instead} add moves directly in the space of permutations.

The simplest strategy consists of swapping two elements sampled uniformly at each step.  It then takes $(n-1)$ steps to build an irreducible chain on the space of permutations.  The size of the neighbourhood is independent of the partition and fixed at $L$.

However, swapping nodes inside the same partition element does not change anything and just increases the probability of staying still.  Excluding such possibilities and only swapping nodes in different partition elements, the neighbourhood is then smaller and equal to
\begin{equation} \nonumber
\sum_{i=1}^{m} \frac{k_i (n-k_i)}{2} \, .
\end{equation}

Finally, one could consider only allowing nodes in \edity{adjacent} partition elements to be swapped.  The idea here would be that this move would be more `local' and be more likely to \corry{choose} a swap with a high score.  Only a smaller number of nodes need to be rescored leading to faster steps in the chain. The neighbourhood is smaller still and equal to
\begin{equation} \nonumber
\sum_{i=1}^{m} k_ik_{i+1} \, ,
\end{equation}
but the irreducibility length in the space of permutations increases up to order $n^2$.  \edity{Accordingly, local steps get stuck more easily in high scoring areas reducing convergence and mixing but as compensation more moves can be performed in the same computational time.  \corry{We} can balance the two types of permutation moves to benefit from the positive aspects of both and improve the overall convergence.  In particular we again weight them} to keep the average number of nodes to rescore down to about 4.

\subsection{Combining moves}

The partition and permutation neighbourhoods could be combined into a single larger neighbourhood to sample uniformly from (including the current point).  For simplicity\corry{,} we sample each move type with a fixed probability.

\section{Partition MCMC with edge reversal}

In the edge reversal move of \cite{gh08}, new parent sets of the two nodes connected by the edge selected for reversal are also sampled.  Since they are sampled according to their score, higher scoring DAGs are proposed more often\corry{,} and the chain moves more quickly.   However, the move is non ergodic\corry{,} and needs to be built on an underlying irreducible framework like structure MCMC, which was the choice in \cite{gh08}.

Since partition MCMC offers an alternative to structure, it can be used as the underlying MCMC method instead.  A DAG is simply sampled from the current labelled partition according to the list of scores, the new edge reversal move from that DAG is performed exactly as in \cite{gh08}\corry{,} and the proposed DAG is accepted with the same probability.  If accepted, the new DAG is mapped to its labelled partition\corry{,} which is then used for further partition MCMC steps.  Since the starting DAG is sampled from the current partition, the relative scores of the start and end partition cancel (like the steps in the new edge reversal move) and we need no further correction.

Explicitly, the transition probability from $\Party$ to $\Party'$ through the DAGs $\graph\in\Party$ and $\graph'\in\Party'$ is
\begin{equation} \label{partedgerevtransitionprob}
K_{\graph' \vert \graph}(\Party' \vert \Party) = \frac{P(\graph\vert \obs)}{P(\Party\vert \obs)} K^{\triangleright} (\graph' \vert \graph) \, ,
\end{equation}
where the first term is the probability of sampling a DAG from the partition with normalisation as in \eref{partitionscoreeq} and $K^{\triangleright} (\graph' \vert \graph)$ is the transition probability of the edge reversal move in \cite{gh08} from $\graph$ to $\graph'$.  Since that move satisfies detailed balance
\begin{equation} \nonumber
\frac{P(\graph'\vert \obs)}{P(\graph\vert \obs)} = \frac{K^{\triangleright} (\graph' \vert \graph)}{K^{\triangleright} (\graph \vert \graph')} \, ,
\end{equation}
substituting into \eref{partedgerevtransitionprob} leads directly to
\begin{equation} \label{partedgerevdetailedbalance}
\frac{K_{\graph' \vert \graph}(\Party' \vert \Party) } {K_{\graph \vert \graph'}(\Party \vert \Party') } = \frac{P(\Party'\vert \obs)}{P(\Party\vert \obs)} \, , 
\end{equation}
so that detailed balance holds for the edge reversal move inside partition MCMC when the DAG is sampled from the current labelled partition.

Finally \corry{we need} to consider the possibility that there is more than one path between partitions so that
\begin{equation} \nonumber
K(\Party' \vert \Party) = \sum_{\graph,\graph'}K_{\graph' \vert \graph}(\Party' \vert \Party) \, ,\end{equation}
is the total transition probability between the two partitions.  Detailed balance
\begin{equation} \nonumber
P(\Party\vert \obs)K(\Party' \vert \Party) =  K(\Party \vert \Party') P(\Party'\vert \obs) \, ,
\end{equation}
follows from \eref{partedgerevdetailedbalance}, rearranged and summed.

The computational expense of an edge reversal move involves examining the possible parent sets four times.  Namely, for both nodes attached to the edge which is reversed and for both the forward and backward move.  It therefore has roughly the same computational cost as a partition MCMC move.  However, if the move is accepted\corry{,} the entire new partition needs to be scored from scratch by rescoring all the nodes.  Therefore a relative cost of approximately $\frac{n}{4}$ is added to each accepted move, slightly reducing the length of the chain compared to standard partition MCMC.

\section{Conclusions}

As demonstrated in \aref{app}, partition MCMC allows DAGs to be sampled from the posterior much more efficiently than standard structure MCMC, without the bias of order MCMC \citep{fk03}.  The current state of the art edge reversal move of \cite{gh08} can be built into the partition sampler\corry{,} improving on the convergence of their algorithm which required an underlying structure MCMC.
Partitions are a way of putting MCMC on DAGs in a more natural mathematical framework, where the algorithm acts on their combinatorial representation.  This representation may provide a means to define distances alternative to the SHD via moves in the space of partitions. As such\corry{,} partition MCMC opens up new possibilities.  Here we chose some simple moves to demonstrate the idea, but many more could be defined and the choice optimised, maybe even adaptively depending on the score landscape.  Structure MCMC on the other hand is a very mature methodology and therefore highly optimised over the years\corry{,} while \corry{it is} hard to envisage simple modifications of the edge reversal move of \cite{gh08}. The combinatorial approach we present, being novel, may offer wider scope for improvement.

\edity{With the aim of inferring a single Bayesian network from scarce data, \cite{elidan11} suggested an adaptation of structural EM based on bootstrap aggregating or bagging \citep{efron79,breiman96}.  The focus here is instead on sampling from the posterior, in order to enable Bayesian model averaging, for example to estimate posterior probabilities of given structural features.  As noted by \cite{fgw99}, an approximation to Bayesian estimation can be obtained from the bootstrap approach.  However, the quality of the estimates decreases when moving away from the mode.  In other words, the structure set obtained from the bootstrap samples provides a distribution of the maximum likelihood estimator and as such does not lead to an adequate description of the posterior.  Typically the maximal likelihood search is performed in the structure space, but a more natural approach would be to use the order or partition space as we discuss in \aref{MAPdiscovery}.  However the complexity of such a search is the same as sampling from the posterior.  Hence the bagging approximation can be avoided since partition MCMC provides access to the full posterior.}

\edity{In \aref{app} we look at example of densely connected graphs with up to 20 nodes.  From the complexity arguments also discussed in \aref{MAPdiscovery}, fixing the computational cost we can increase the number of nodes while decreasing the limit on the number of parents.  For example, graphs with 20 nodes and no parent limit would be comparable to graphs with 100 nodes with up to 2 or 3 parents each.}

In some cases some of the structural features of the domain under study may be known \corry{from} previous studies or expert knowledge. It is then useful to include prior information in the learning process, as suggested for example by \cite{ms08, wh07}. Modular priors, such as on edges\corry{,} can be easily accounted for in partition MCMC, order or edge reversal. Non modular priors could be included in structure MCMC \citep{ms08}, but the slow convergence of the algorithm makes this of little practical interest for domains of moderate size, therefore they would most naturally be corrected for by means of importance sampling.

The idea behind partition MCMC is analogous to order MCMC which may be seen as an elegant solution to the \corry{somewhat different} problem of sampling triangular matrices.  By simply enforcing the chain to respect the combinatorial structure of DAGs via the partitions\corry{,} we can now solve the problem of interest.  Alternatively, one could wish to sample CPDAGs, in which case partition MCMC would still be solving slightly the wrong problem.  \cite{madiganetal96} proposed a method to sample essential graphs (or CPDAGs) as an extension of structure MCMC, which can be sped \corry{up} following the approach of \cite{pena07} for the uniform case.  \edity{Recent improvements also allow the uniform sampling of large sparse essential graphs \citep{hjy13}} However, \blindey{as discussed by \cite{km13}, }the convergence on CPDAGs is notably slower than for DAGs\corry{,} and at present no better methods like edge reversal \citep{gh08} or partition MCMC are available for the space of CPDAGs.  Moreover\corry{,} the overcounting of CPDAGs when working on DAGs instead is bound by a low constant and is approximately 4\corry{,} so that rejection or importance sampling should be preferable when combined with an efficient method for DAGs.  \edity{For example one could try to approximately evaluate feature prevalence in the essential graph space following the ideas of \cite{ew08}.  Namely, assuming each essential graph has a different score, and since equivalent DAGs necessarily have the same score, one would only keep a single copy of any equally scoring DAGs.}

In contrast the overcounting of lower triangular matrices compared to DAGs grows exponentially\corry{,} explaining the difficulties encountered when employing importance sampling to correct for the bias of order MCMC.  With an implicit assumption that such a correction via importance sampling is possible, there has been work to improve order MCMC by working on partial orders \citep{npk11} or sampling directly on the space of orders using dynamic programming \citep{ks04}.  Currently\corry{,} an improved dynamic programming approach to order MCMC has been proposed \citep{htw15} with approximate bias removal following the work of \cite{ew08}.  Of course\corry{,} none of these approaches can sample DAGs from the posterior as with our partition MCMC approach.  \corry{However} trying to build the dynamic programming framework on the space of partitions rather than orders may be an interesting direction to explore as a possible way to sample DAGs correctly and efficiently.


\appendix
\section{Comparison of the different MCMC methods}
\label{app}

The standard set of moves through the space of DAGs is to change one edge at a time \citep{my95,gc03}. Which edges can be added, deleted or reversed can be calculated with the help of the incidence and ancestor matrices. 
For example each nonzero entry in the incidence matrix can be set to 0 to delete an edge, giving the neighbouring graphs with one edge fewer.  Edges cannot be added to ancestors\corry{,} as this would create a cycle, or to where an edge already exists.  The neighbours with one edge more can thus be easily found.  Finally, edge reversal can be thought of as a particular two step move of deleting an edge and adding it back in the opposite direction.  Appropriately multiplying the ancestor and incidence matrices, provides the nondirect ancestors.  Reversing the edge to any of them would create a cycle. The neighbours with an edge reversed are then derived from the remaining nonzero entries of the incidence matrix.

Given the ancestor and incidence matrices, calculating the neighbourhood is $O(n^2)$ for the edge additions and $O(n^3)$ for the edge reversals. (We ignore possible faster matrix multiplication algorithms.)  Once a new DAG is sampled from the neighbourhood, the ancestor matrix can be updated in $O(n^2)$ \citep{gc03} and the new neighbourhood size needs to be calculated before accepting the move.

We work with the BGe score \citep{gh94,hg95,gh02}, corrected as in \cite{cr12,kmh14} and sped up following \cite{kmh14}.  Since this score is modular, each node is scored separately from the others just depending on its parent set.  With edge addition or deletion, only one node needs to be rescored while two are rescored with an edge reversal.   The scoring involves finding the determinant of a matrix of the size of the parent set with a complexity up to $K^3$ when the size of the parents sets are limited to $K$.

Since here we limit the parent sets, the scoring may be quicker than calculating the neighbourhoods and there have been proposals to improve the speed of the chain by avoiding to explicitly calculate the neighbourhood at each step. 
In an approach designed for uniform sampling \citep{mdb01} one can simply swap an element (from 0 to 1 or back) of the incidence matrix chosen uniformly. 
Moves which would create a cycle are rejected and can be checked in a time between $O(n)$ and $O(n^2)$ with a typical $O(n\log(n))$ behaviour \citep{ar78} which here would depend on the higher scoring DAGs. 
A reversal move can also be introduced \citep{mp04}.  Alternatively, \cite{gc03} keep track of the ancestor matrix which allows them to check whether the moves are legal in $O(n)$. 
Only when moves are accepted based on the score does the ancestor matrix need to be updated at a cost $O(n^2)$. Of course convergence is slowed down by not calculating the neighbourhood exactly but the idea is that this is more than compensated for by speeding up the moves.

Since the exact computation time depends on many factors we employ the following simplifications in our comparison. We find the number of steps of standard structure MCMC that have approximately the same computational time as the alternatives detailed below and then run the chains $n$ times longer.   Although the examples are far from the asymptotic limit where edge reversal is more expensive than the other moves, this factor is chosen is to compensate for possible speed ups that could be implemented. With the compensation, the unavoidable time spent scoring DAGs is comparable to the run time of the alternatives, representing the limit of any speed ups.

Both \cite{fk03} and \cite{gh08} suggested a more constant time factor $\approx 10$ between structure and alternative steps.   Although it may depend heavily on the exact implementation, it wouldn't be in line with asymptotic reasoning but for $n\approx 10$ would be roughly in line with our implementation. Our comparison however would become more favourable to structure MCMC as $n$ increases.

The standard move in order \corry{MCMC} consists of swapping any two elements in the permutation of node labels (so the chain takes less than $n$ steps to become irreducible).  However, it means you need to rescore the nodes selected and all the nodes inbetween them (on average $\frac{n+4}{3}$ nodes). Instead one could swap \edity{adjacent} elements and only need to rescore 2 nodes (making the steps quicker) but the chain takes order $n^2$ steps to become irreducible.  We employ a mixture of both moves with the probability of each chosen so that they both take the same computational time.  Since the standard move includes the other, it should be chosen with probability $6n/(n^2+10n-24)$ to achieve this, meaning that around 4 nodes need to be rescored on average.  This improves the behaviour of order MCMC and makes the time complexity of each step independent of $n$.  A small 1\% probability of staying still is included to ensure aperiodicity of the chain.

For partition MCMC we choose a partition move to a permutation move with a ratio of $3:2$ and inside each class we choose the larger move with $6n/(n^2+10n-24)$ so that on average about 3 nodes and one partition element (typically a further node) are rescored at each step.  Likewise a small 1\% probability of staying still is included.

The edge reversal move of \cite{gh08} is not irreducible by itself and it needs to be incorporated into an irreducible scheme like structure MCMC.  Since for the edge reversal move we need to score 4 lots of possible parent sets, 2 for the forward move and 2 for the reverse move, this takes a similar time on average to each partition MCMC move.  For the comparison we fix the probability of the new edge reversal move to $7/100 \approx 1/15$ as in \cite{gh08} and use the artificial timing of the structure steps to set the total time to match partition MCMC.  Note that the edge reversal move requires knowledge of the descendants matrix once the edge to be reversed has been removed.  This is obtained in a naive matrix multiplication implementation of order $n^4$ compared to the order $n^K$ of scoring and sampling parent sets.  Despite the complexity, the actual computational time was negligible in our examples and faster than a less complex algorithm, but this step could be sped up for larger graphs.

When the edge reversal move of \cite{gh08} is combined with partition MCMC instead we keep the probability of choosing the edge reversal move \corry{at} $7/100$.  If such a move is accepted though we need to rescore all the nodes for the next partition step which slows down the implementation and leads to marginally shorter chains.

\subsection{Simulated example}

We generated 100 observations from the DAG in \fref{dagexample} with 5 nodes and no maximum number of parents ($K=4$).  The data were generated from a normal with regression on the parents with coefficient $2$.

\begin{figure}
  \centering
$\begin{array}{cc} \includegraphics[width=0.45\textwidth]{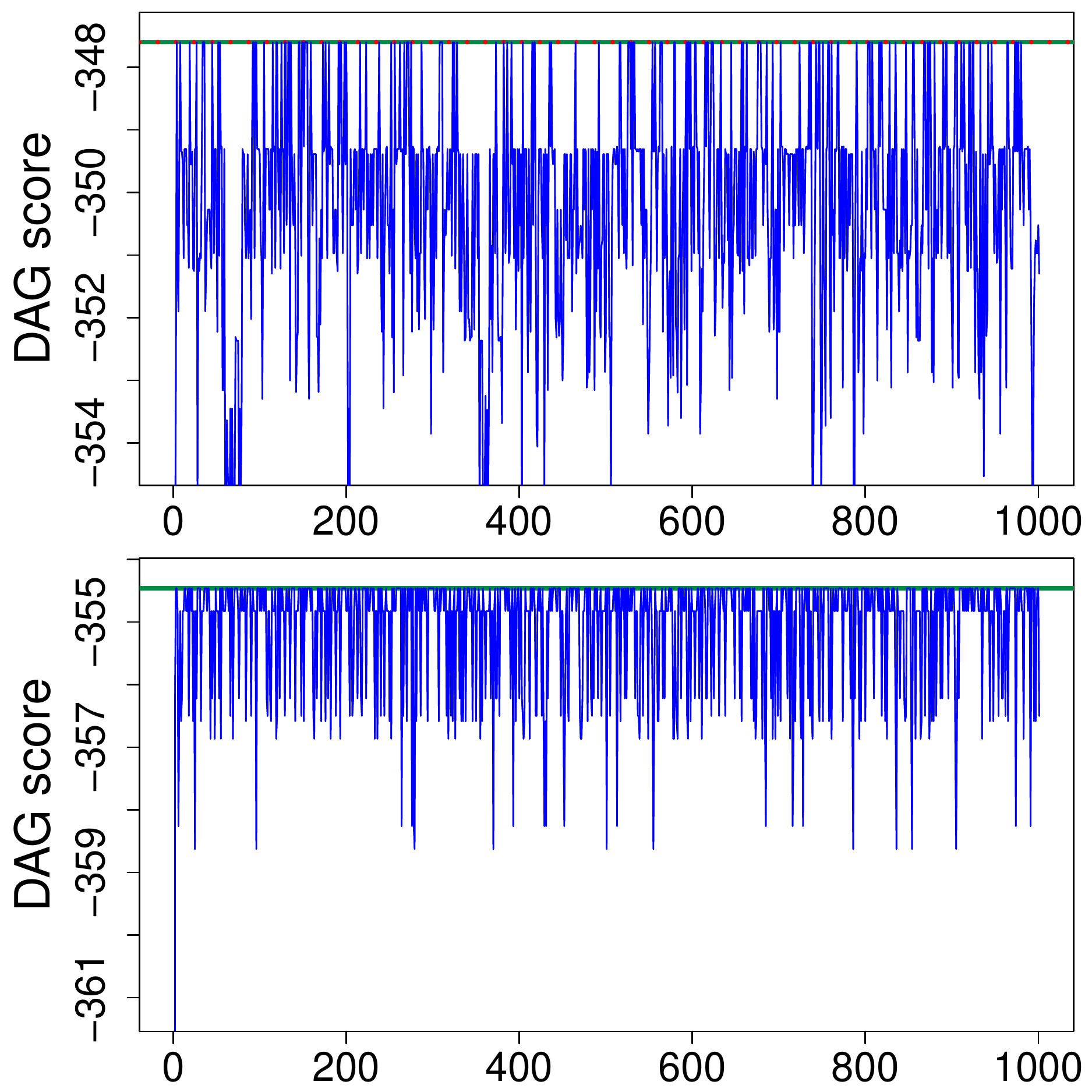} & 
 \includegraphics[width=0.45\textwidth]{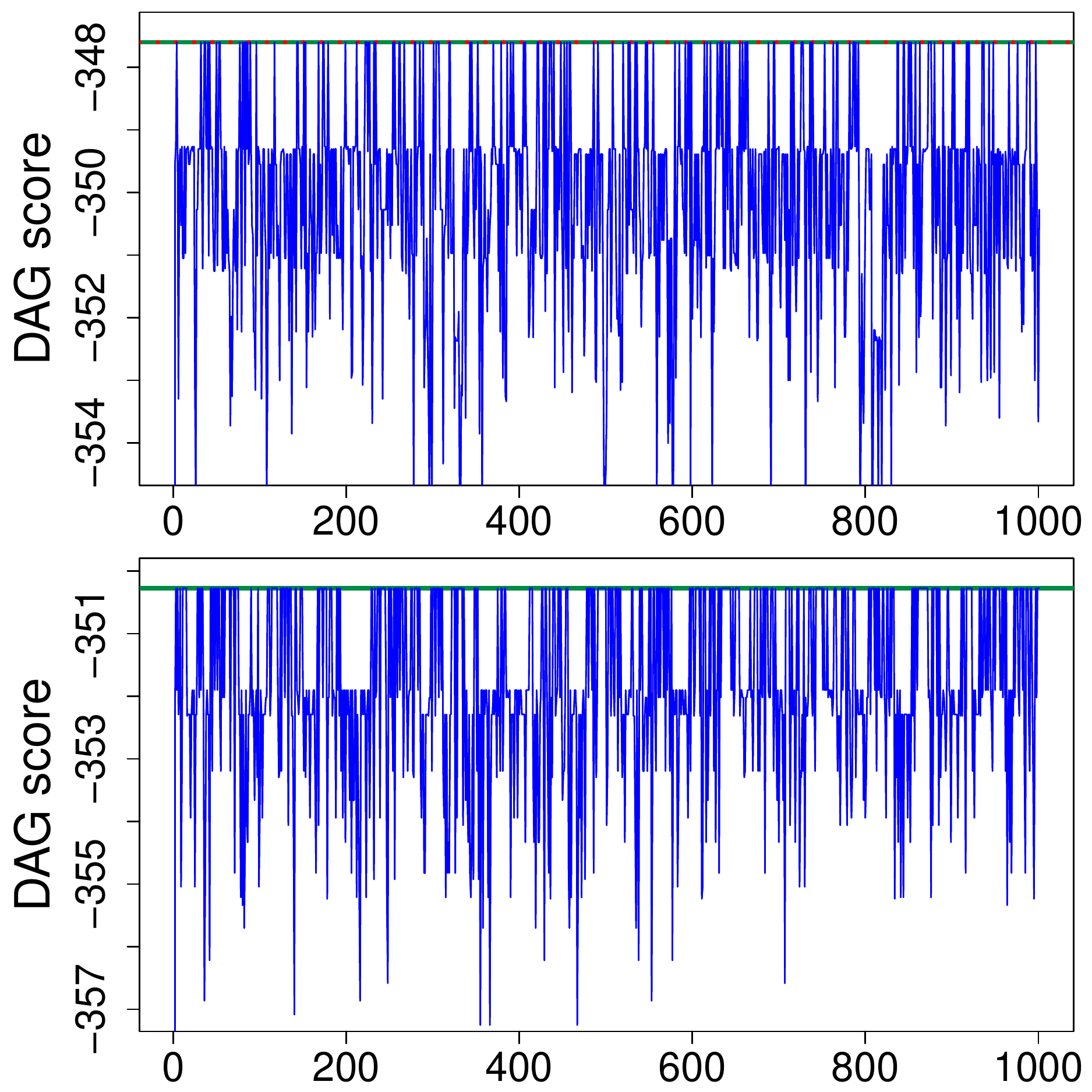}
 \end{array}$
  \caption{A run of 50 thousand steps of structure MCMC for the simulated data with different seeds starting at the empty DAG.  In the top plots edge reversals are allowed and we seem to hover around the maximally scoring DAG (red dotted line).  In the bottom plots, edge reversal is excluded and we do not even approach the maximal scoring DAG.}
  \label{structure5nodes}
\end{figure}

First we ran standard structure MCMC \citep{my95} for 50 thousand steps recording a thousand evenly spaced DAGs. \Fref{structure5nodes} shows trace plots from two different seeds.  
The highest score of the DAGs covered in the chain is plotted as the green solid line which is placed at the top of each graph.  If visible, the score of the DAG used to generate the data is plotted as the red dotted line. In the lower plots we excluded the possibility of edge reversal. 
Although, as discussed above, one can speed up the chain without edge reversal, as can be seen in the plots the poor performance of the chain makes this a false economy since the level of the best DAG is not reached. 
The large improvement with edge reversal is detailed in \cite{gc03} and in the following subsections we no longer consider the case without it.

\begin{figure}
  \centering
$\begin{array}{cc} \includegraphics[width=0.45\textwidth]{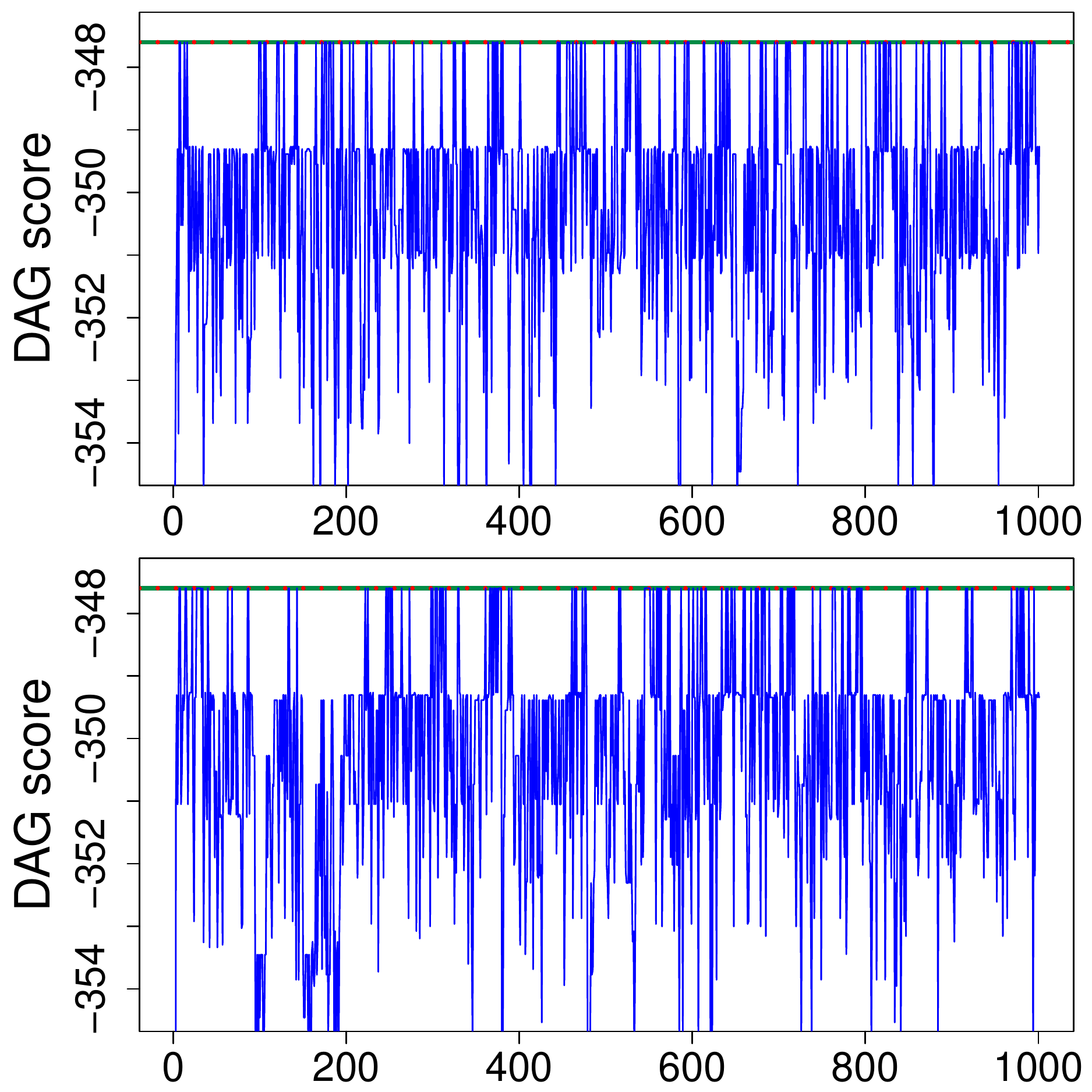} & 
 \includegraphics[width=0.45\textwidth]{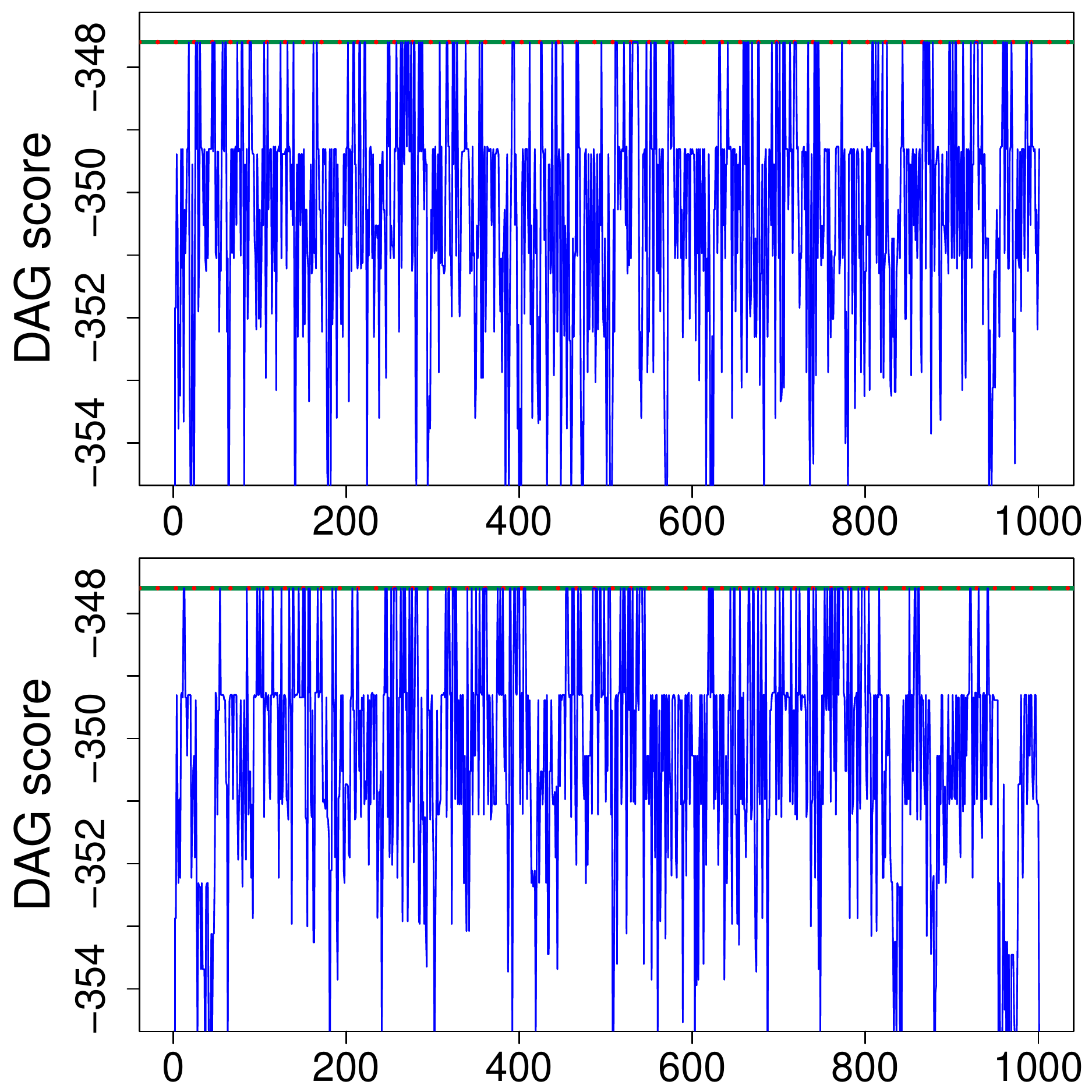}
 \end{array}$
  \caption{A run of 40 thousand steps of structure MCMC with the new edge reversal move of \cite{gh08} and different seeds.  The edge move is selected with probability $0.07$.  The performance seems the same as standard structure MCMC, but with a large improvement when standard edge reversals are excluded as shown by the bottom rows.}
  \label{edgerev5nodes}
\end{figure}

When including the edge reversal move of \cite{gh08} we run the chain for 40 thousand steps.  In this example we ran structure MCMC with and without the standard edge reversal move.  In the top plots of \fref{edgerev5nodes} there is hardly any difference from those in \fref{structure5nodes} other than maybe a slight worsening due to a shorter chain.  The new edge reversal vastly improves the algorithm without a standard edge reversal, as can be seen by comparing the bottom plots, but this is not surprising given the addition of some type of edge reversal.

Removing the standard edge reversal could lead to a speed up of structure, but a degradation in the performance is still evident when comparing the two rows of \fref{edgerev5nodes}.  Since the new edge reversal constitutes the computationally most expensive part, any speed up of the structure part would be damped and unlikely to be worth considering in general situations.

\begin{figure}
  \centering
$\begin{array}{cc} \includegraphics[width=0.45\textwidth]{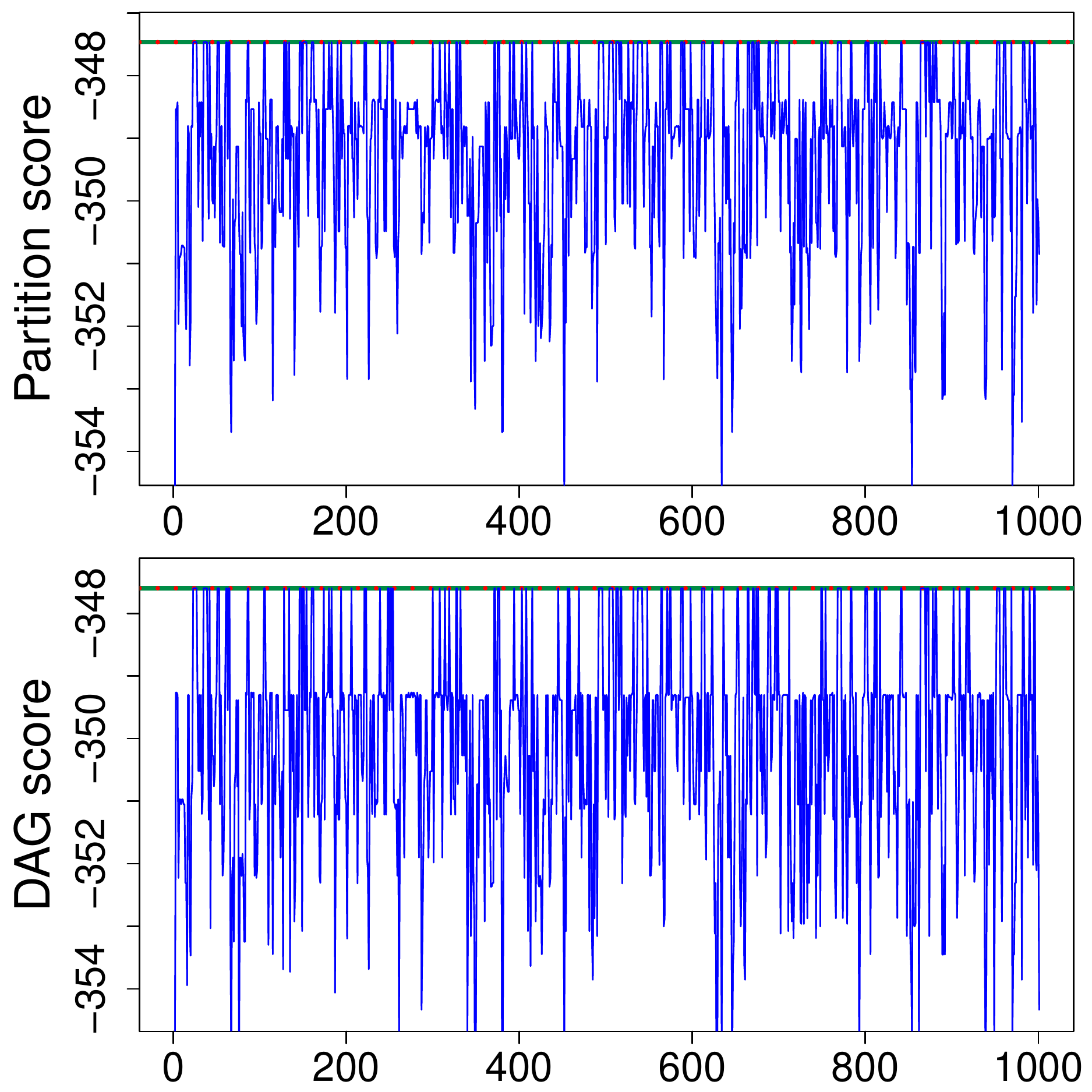} & 
 \includegraphics[width=0.45\textwidth]{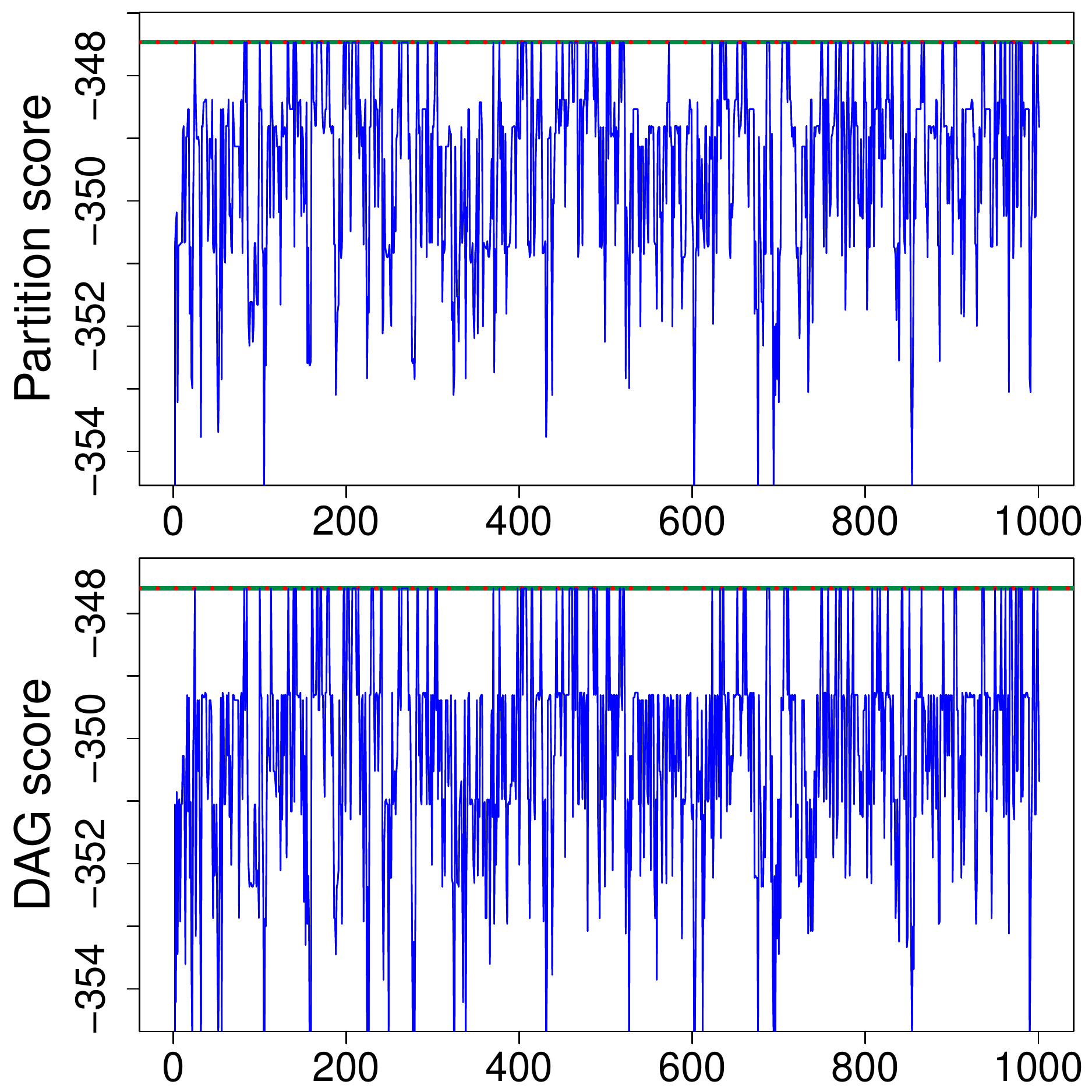}
 \end{array}$
  \caption{Run of 10 thousand steps of partition MCMC with different seeds on simulated data from a DAG with 5 nodes.}
  \label{partition5nodes}
\end{figure}

For the same simulated data, we ran partition MCMC. The time needed for 13 thousand steps (dividing by 5) of structure MCMC allows for approximately 10 thousand steps of partition MCMC.  Some trace plots are given in \fref{partition5nodes} where the top plots show the score of the current partition, while the bottom plots show the score of a DAG sampled from that partition.  When comparing the bottom plots to the top plots of \fref{structure5nodes} we see similar behaviour, with what looks like slower convergence of partition due to the shorter chain.

\begin{figure}
  \centering
$\begin{array}{cc} \includegraphics[width=0.45\textwidth]{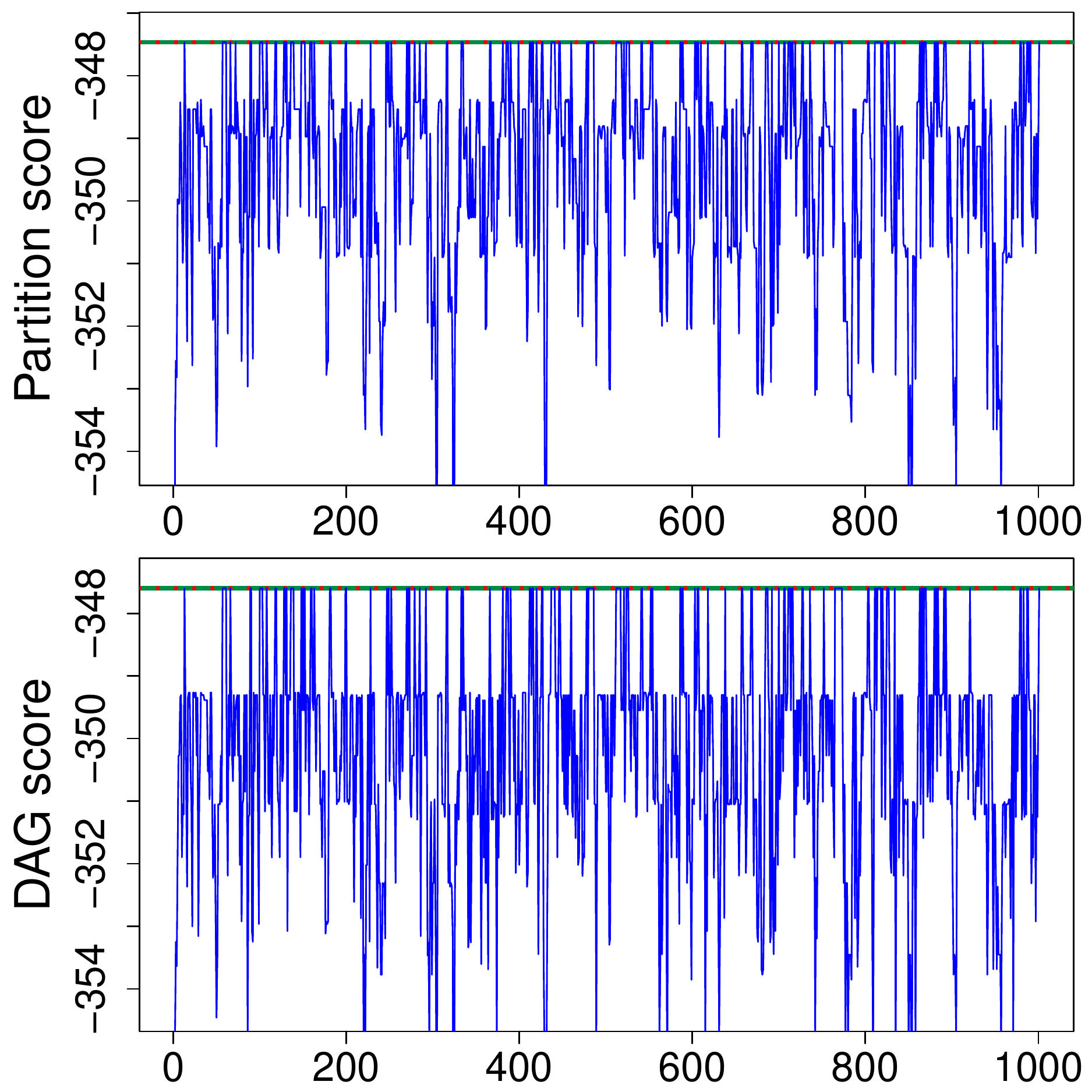} & 
 \includegraphics[width=0.45\textwidth]{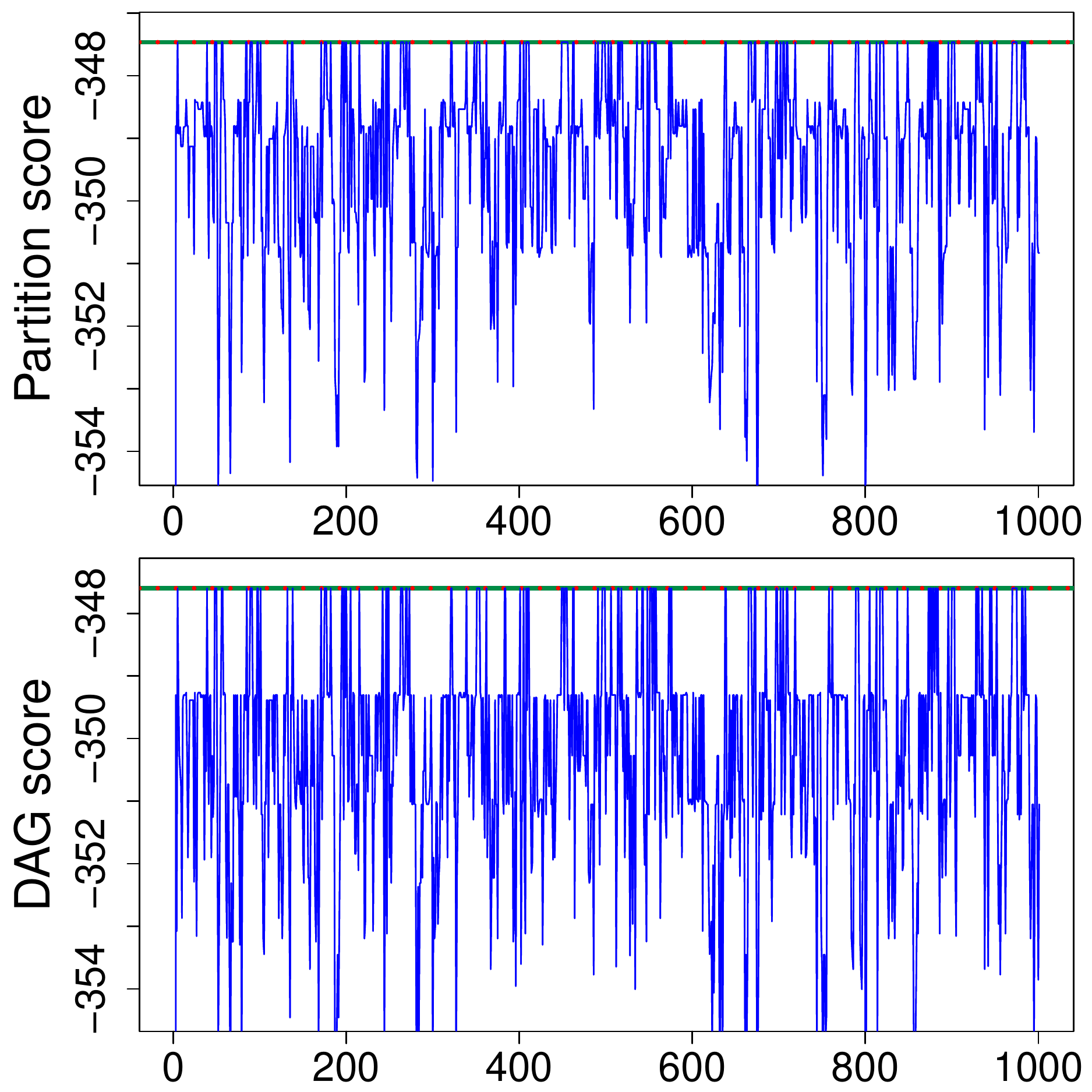}
 \end{array}$
  \caption{Run of 9 thousand steps of partition MCMC including the edge reversal move of \cite{gh08} with different seeds on simulated data from a DAG with 5 nodes.}
  \label{partitionedgerev5nodes}
\end{figure}

Incorporating the new edge reversal move, we can run around 9 thousand steps in the same time with the results in \fref{partitionedgerev5nodes}, which are similar.

\begin{figure}
  \centering
$\begin{array}{cc} \includegraphics[width=0.45\textwidth]{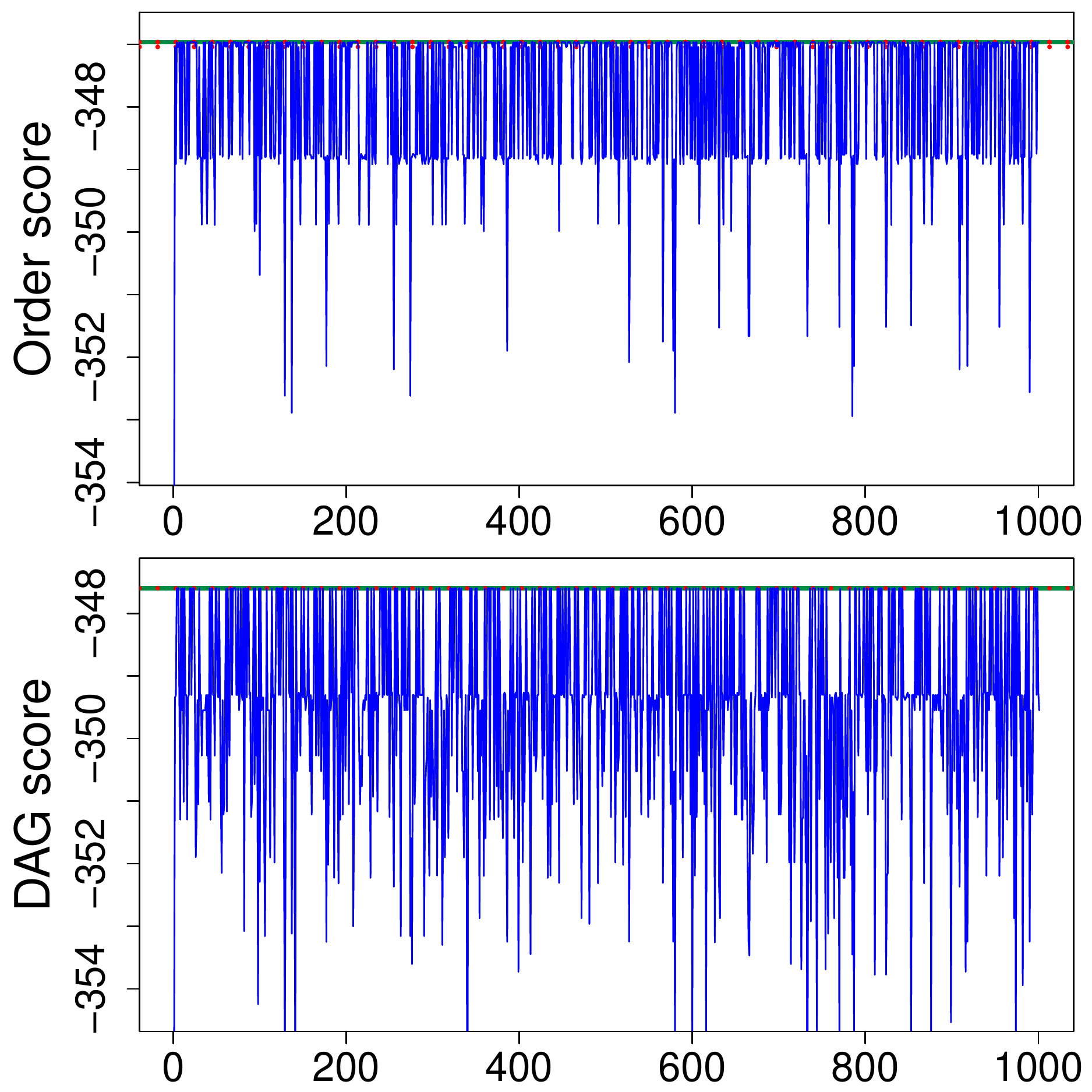} & 
 \includegraphics[width=0.45\textwidth]{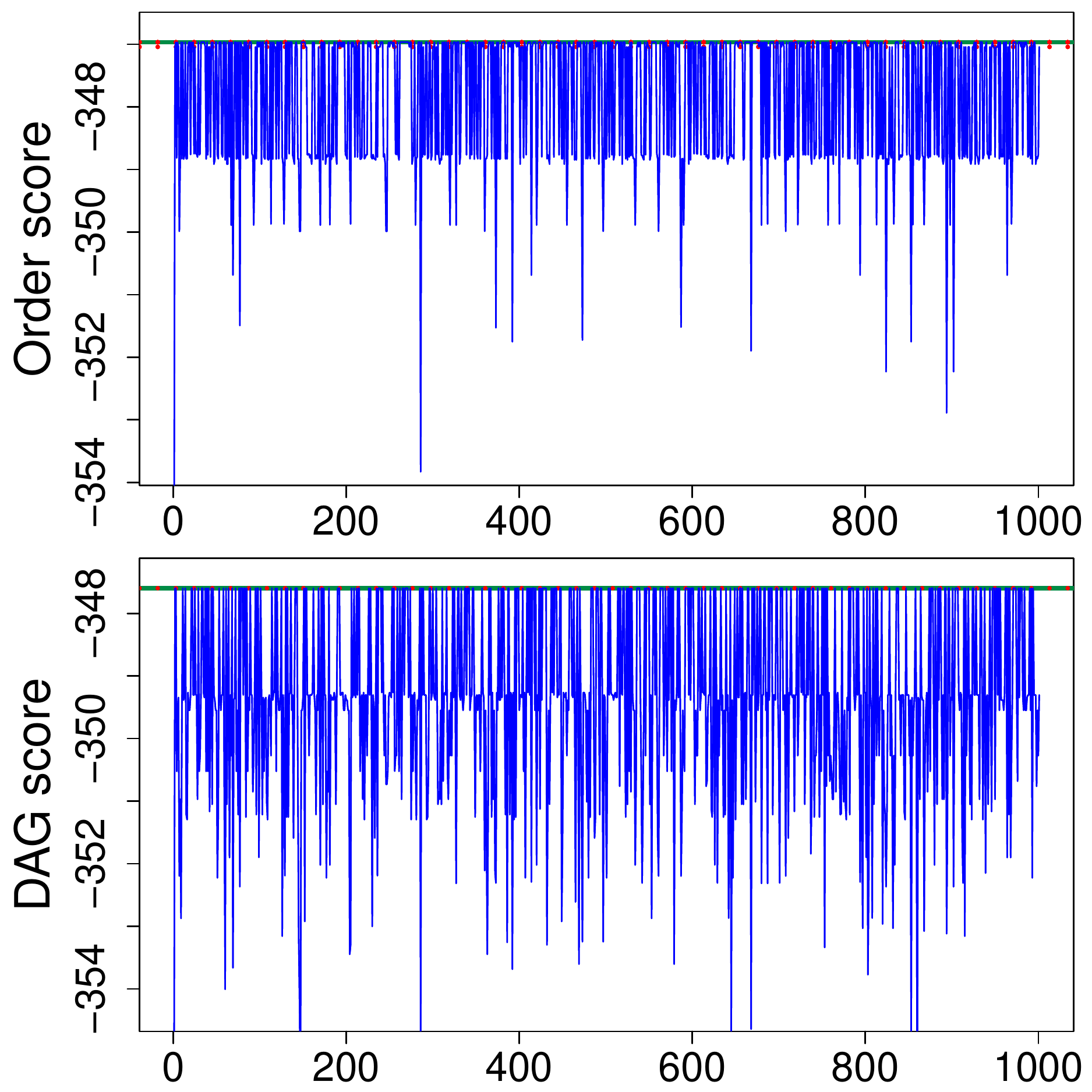}
 \end{array}$
  \caption{Run of 20 thousand steps of order MCMC with different seeds for simulated data from a DAG with 5 nodes.}
  \label{order5nodes}
\end{figure}

Finally we run order MCMC.   In the plots in \fref{order5nodes}, the top line is in the space of orders with the scores of the entire order while the bottom lines are the score of DAGs sampled from each order at each step.  There are now 8 orders the DAG in \fref{dagexample} is compatible with and we draw red dotted lines for the total scores of each of those orders. The performance is much better than structure even with the edge reversal move of \cite{gh08}.  Of course order MCMC is working in a much bigger and overlapping space so we would expect better performance.

\subsection{Boston housing data} \label{BHexample}

A dataset used as a benchmark is the Boston Housing data from the UCI repository \citep{lichman13}.  It consists of $N=506$ observations from $n=14$ continuous variables.  Both \cite{fk03} and \cite{gh08} run their algorithms on the Boston Housing data in order to analyse convergence with respect to structure MCMC.  Here we also use the same dataset to compare the performance of the different algorithms we consider.  We cannot compare directly to the results in \cite{fk03,gh08} since their implementation is not available and moreover they are most likely based on one of the incorrect versions of the BGe score \citep{gh94,hg95,gh02}, which has only recently been corrected \citep{cr12,kmh14}.

\begin{figure}
  \centering
$\begin{array}{cc} \includegraphics[width=0.45\textwidth]{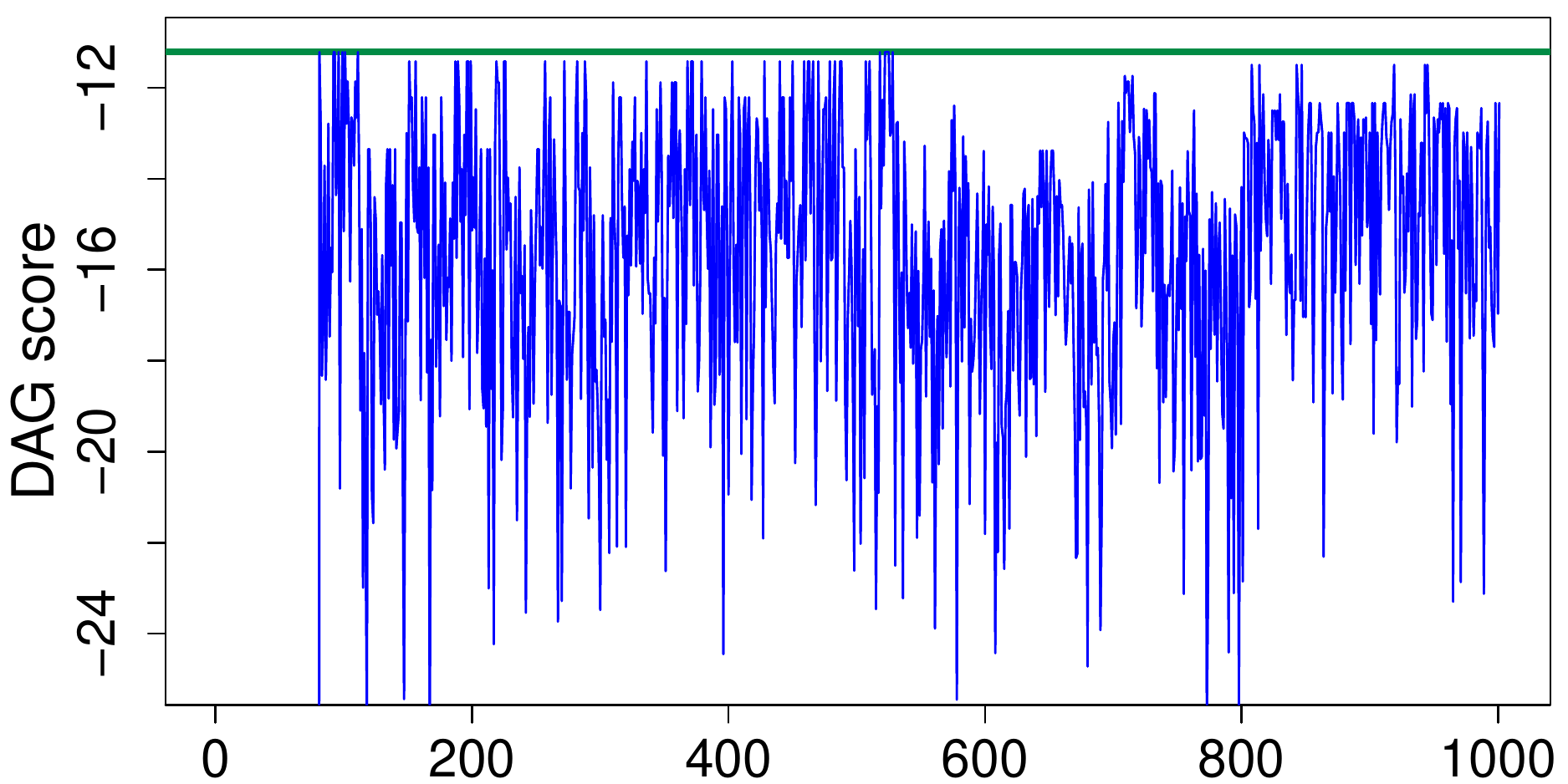} & 
 \includegraphics[width=0.45\textwidth]{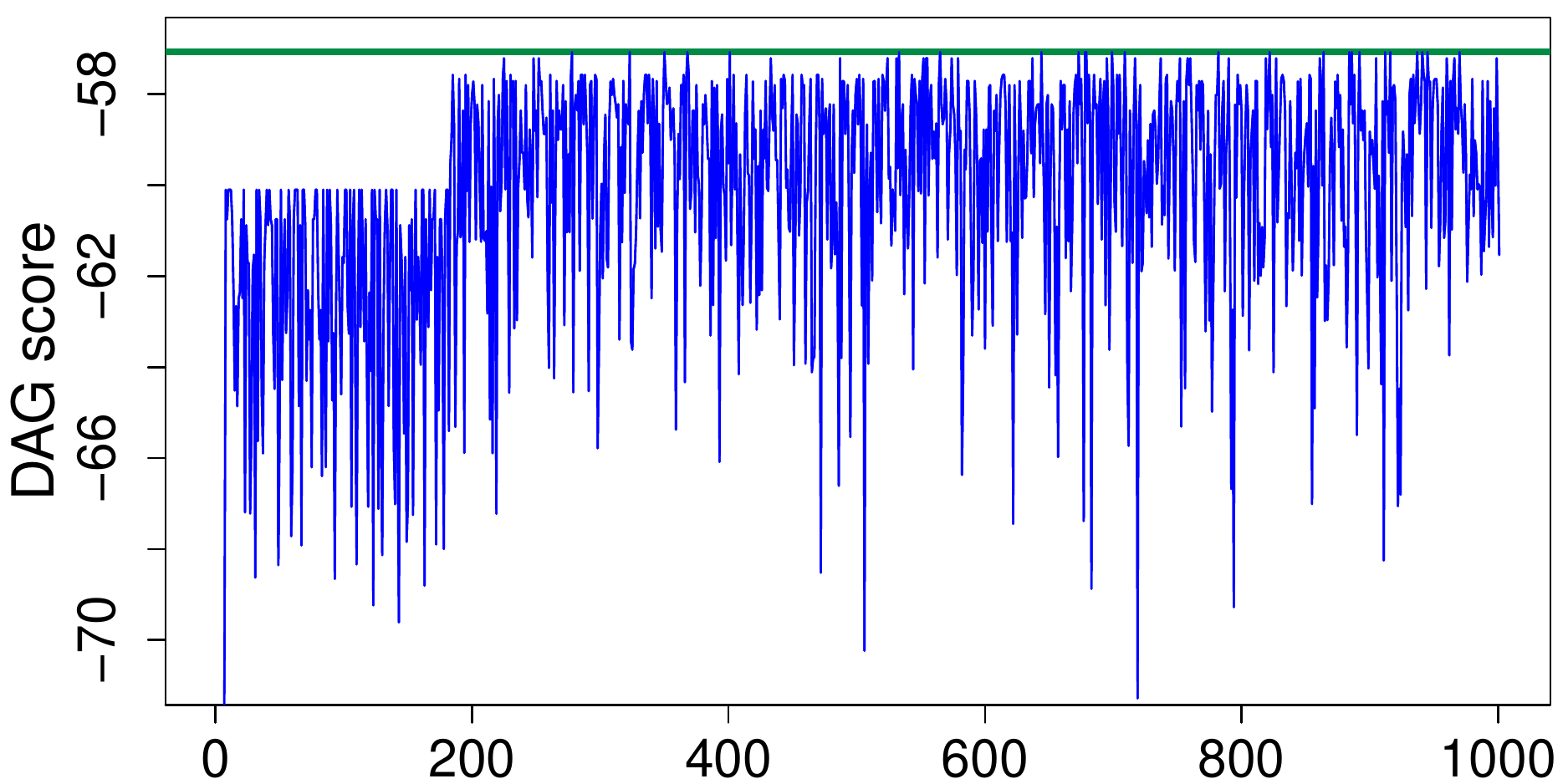} \\
\includegraphics[width=0.45\textwidth]{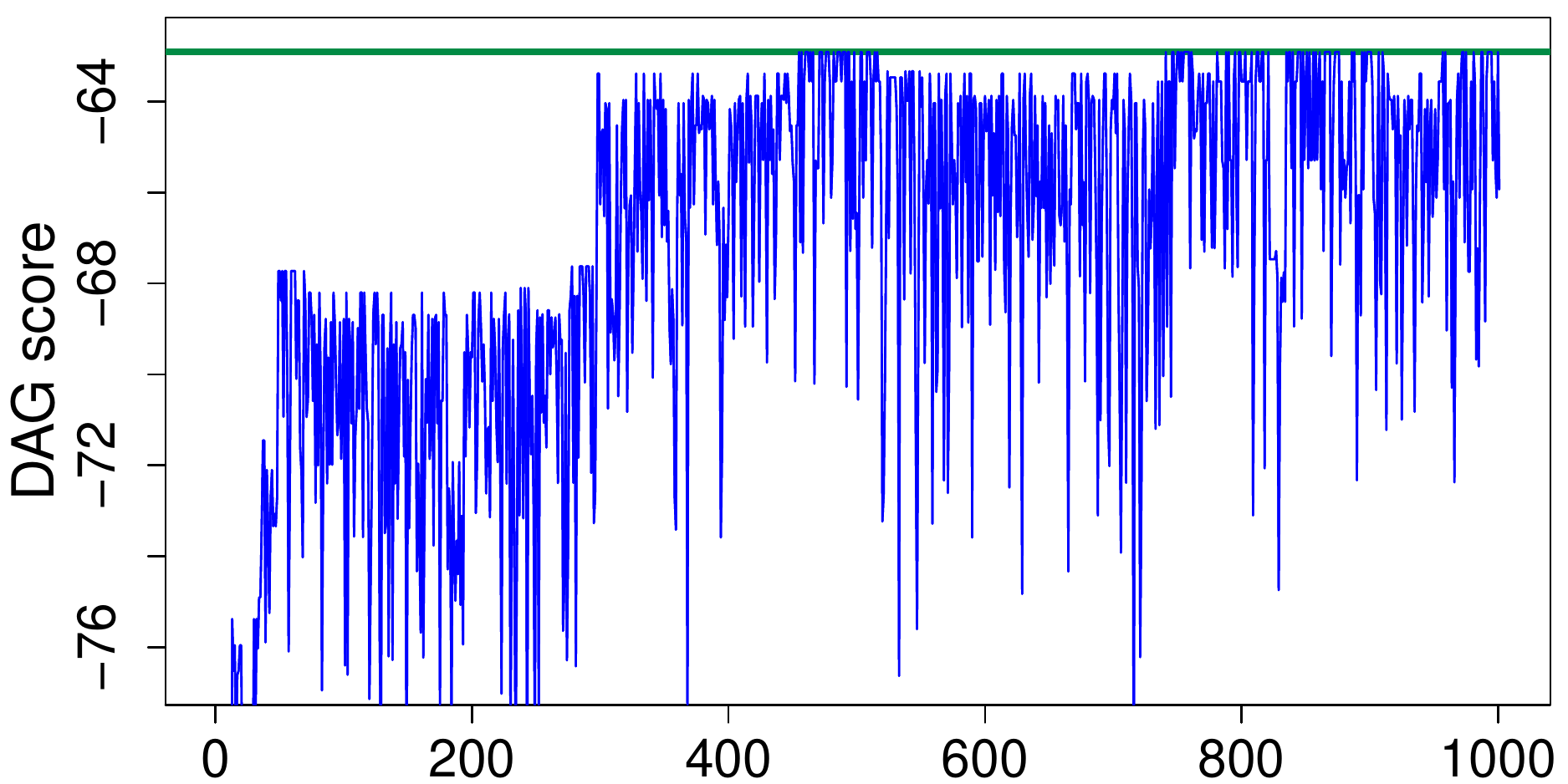} & 
 \includegraphics[width=0.45\textwidth]{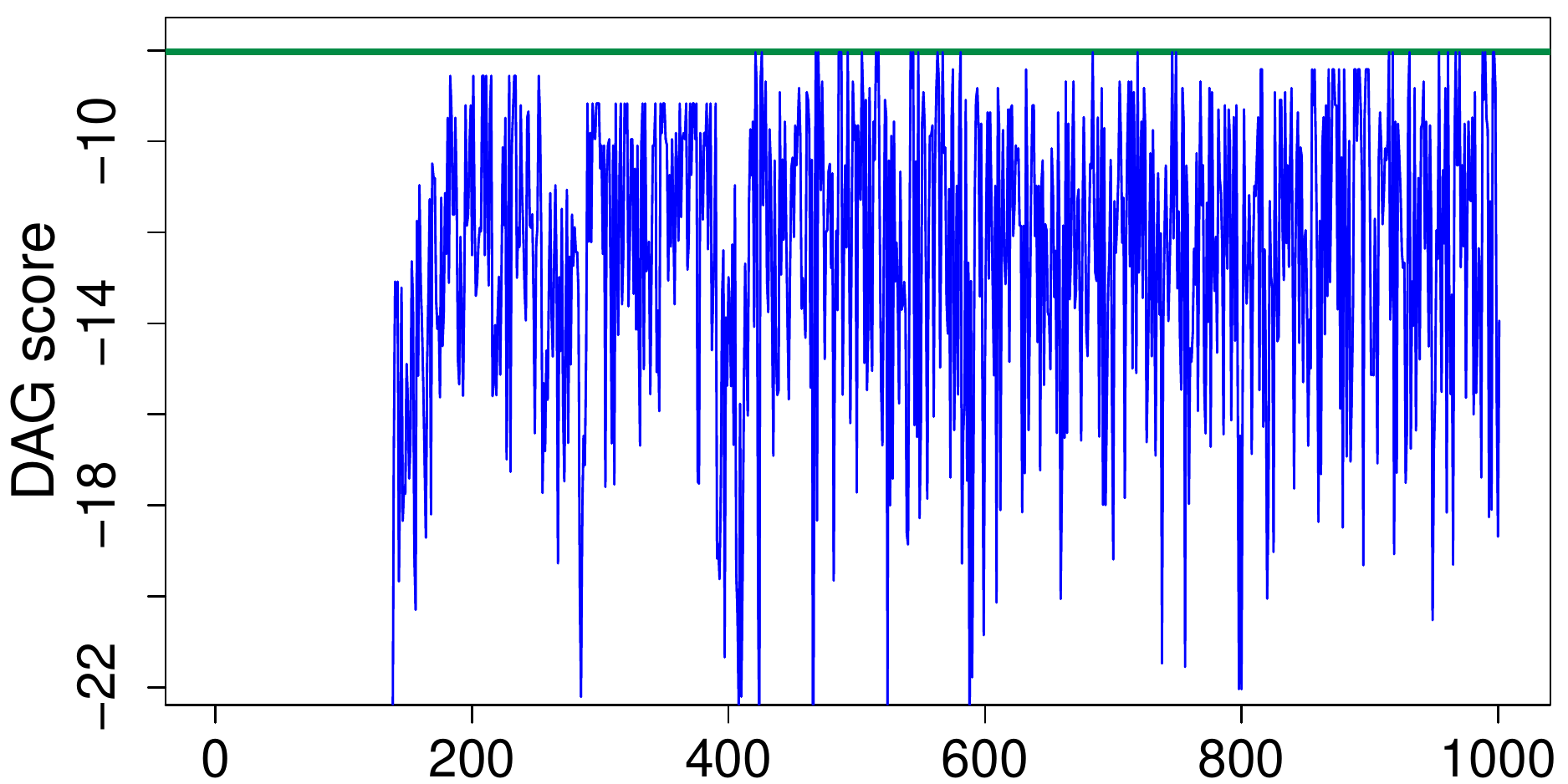}
 \end{array}$
  \caption{A run of 1 million steps of structure MCMC with different seeds for the Boston Housing data.  As can be seen the different runs arrive at different plateaux, some of which are very far away from the global maximum set at 0.}
  \label{structureBH}
\end{figure}

We first run one million steps of structure MCMC with different seeds.  Trace plots are shown in \fref{structureBH}.  Even with such a long run, the chains have not converged.

\begin{figure}
  \centering
$\begin{array}{cc} \includegraphics[width=0.45\textwidth]{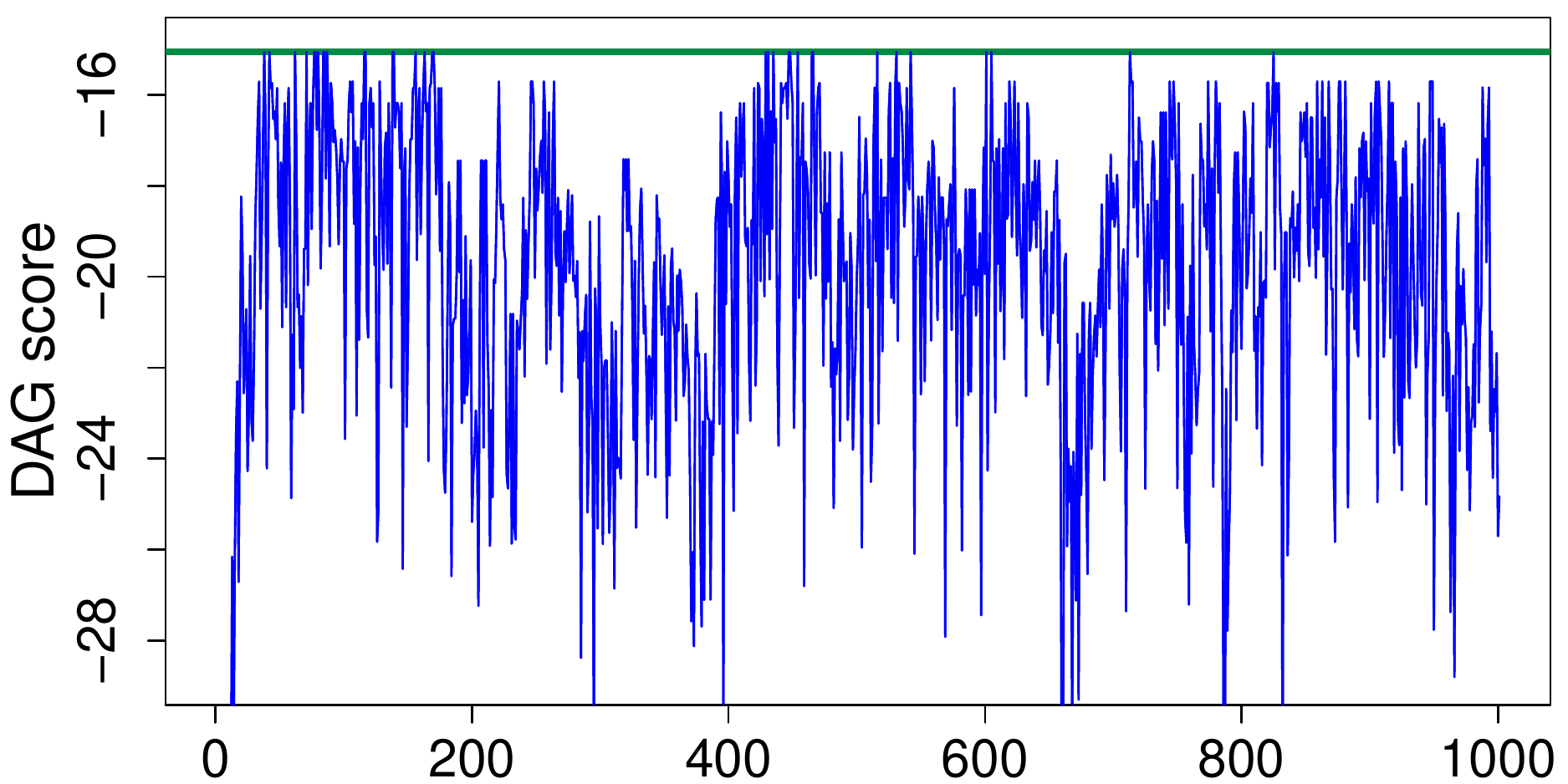} & 
 \includegraphics[width=0.45\textwidth]{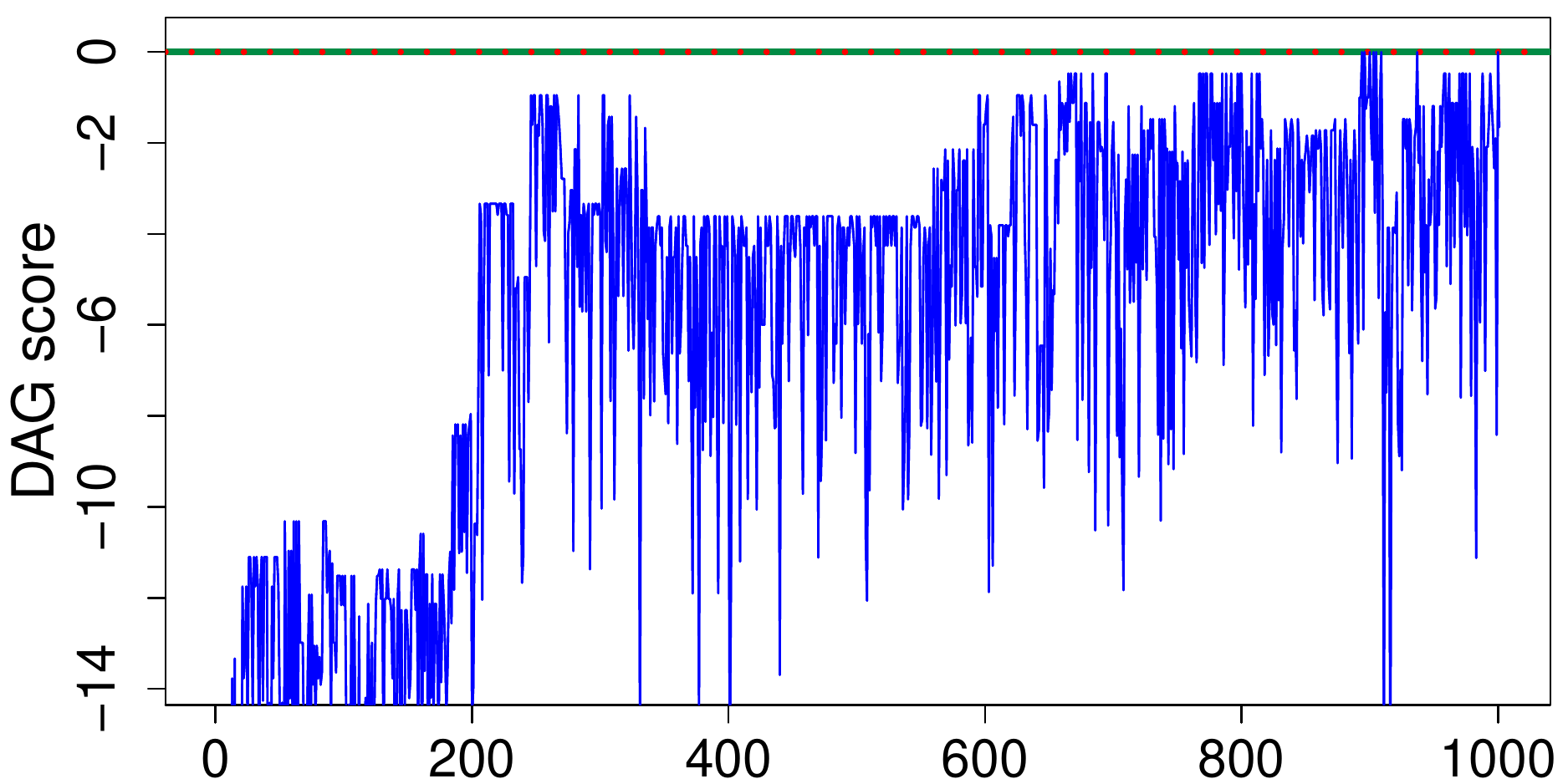} \\
\includegraphics[width=0.45\textwidth]{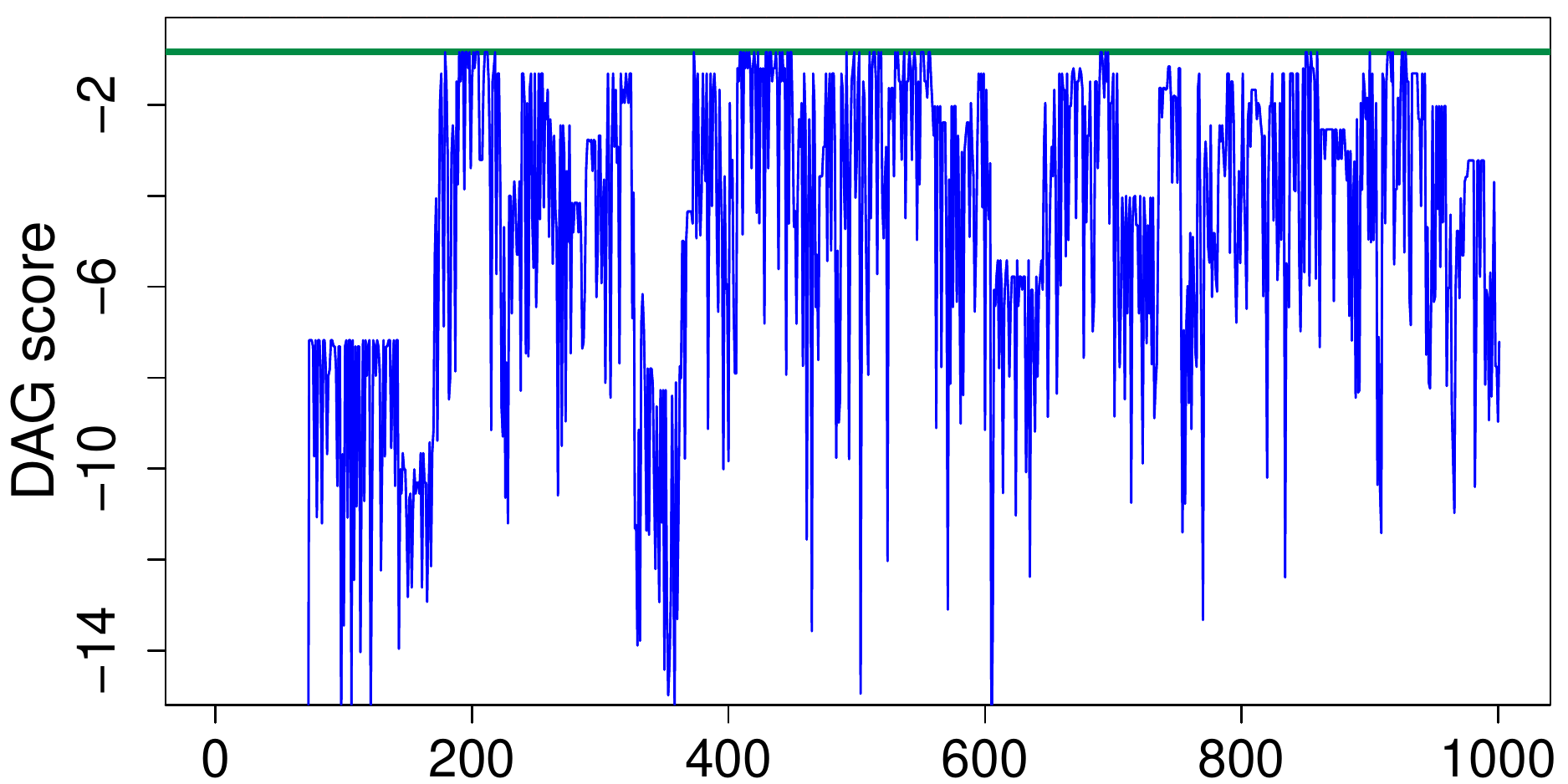} & 
 \includegraphics[width=0.45\textwidth]{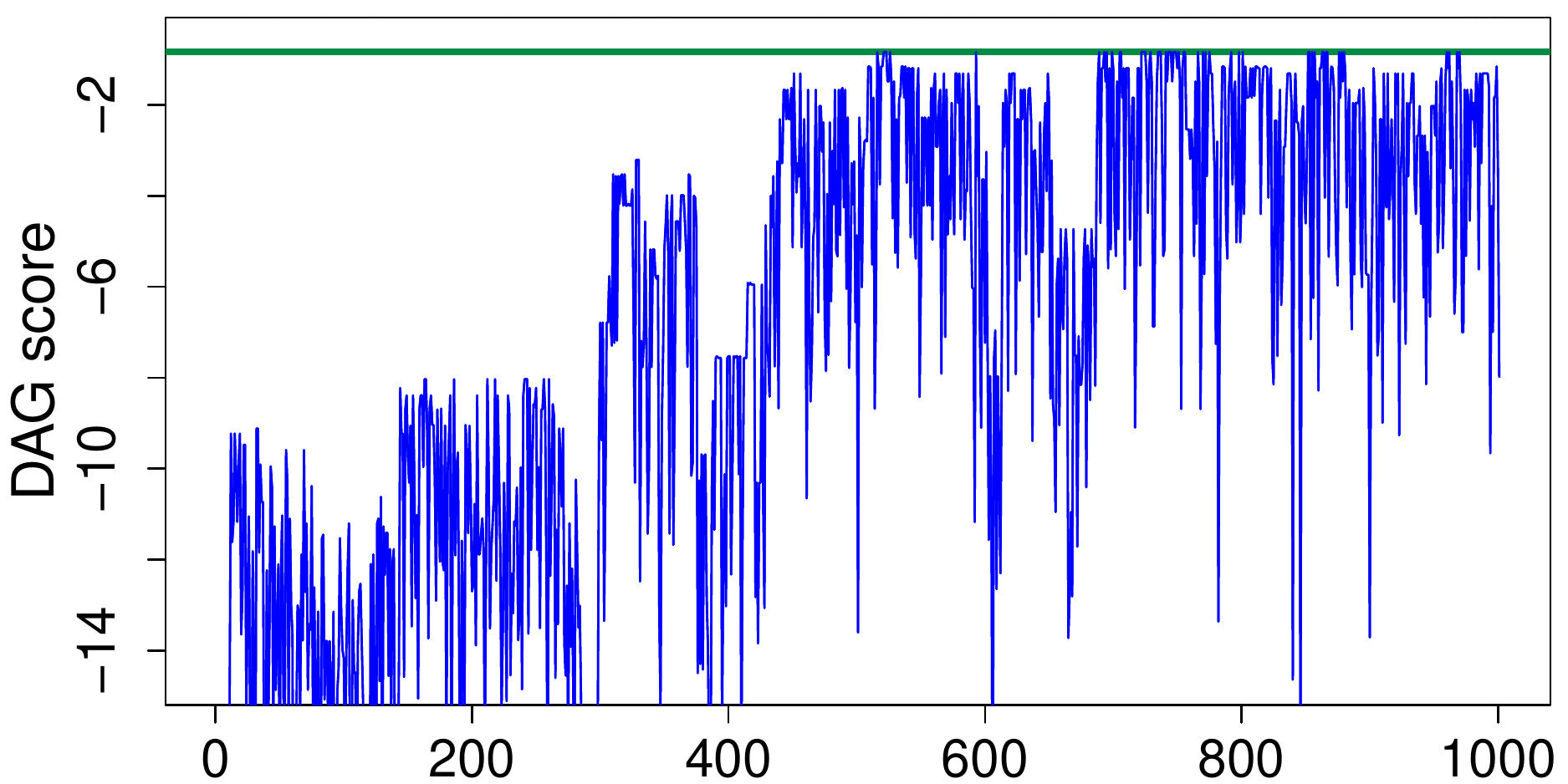}
 \end{array}$
  \caption{A run of 60 thousand steps of partition MCMC with different seeds for the Boston Housing data.  While the first remains some distance away from the global maximum set at 0, the other runs approach a plateau near or at the maximum itself.}
  \label{partitionBH}
\end{figure}

When implementing partition MCMC, dividing by $n=14$ and adjusting the times, we run the chain for 60 thousand steps as opposed to the 1 million we ran for structure MCMC.  Since each partition MCMC step takes a very similar time to an order MCMC step or the edge reversal move of \cite{gh08} we are penalising partition MCMC by a factor of $n=14$ which is comparable to the constant factor of 10 employed by \cite{fk03,gh08}.  When looking at the trace plots in \fref{partitionBH} we can see much better performance than observed for structure MCMC in \fref{structureBH}.  Most of the runs seem to arrive close to where the global maximum resides, which is set to 0 in the graphs.  The remaining run and the deviations from the plateaux give us an idea of the convergence time.

Convergence of structure MCMC is well known to become rather slow as the graphs get larger and the score landscape more peaked.  Partition MCMC seems a simple method to improve the convergence by combining many DAGs in the space of labelled partitions.  Previous approaches to improve structure MCMC such as order MCMC \citep{fk03} and the edge reversal of \cite{gh08} also relied on the combination of DAGs into larger classes.   

\begin{figure}
  \centering
$\begin{array}{cc} \includegraphics[width=0.45\textwidth]{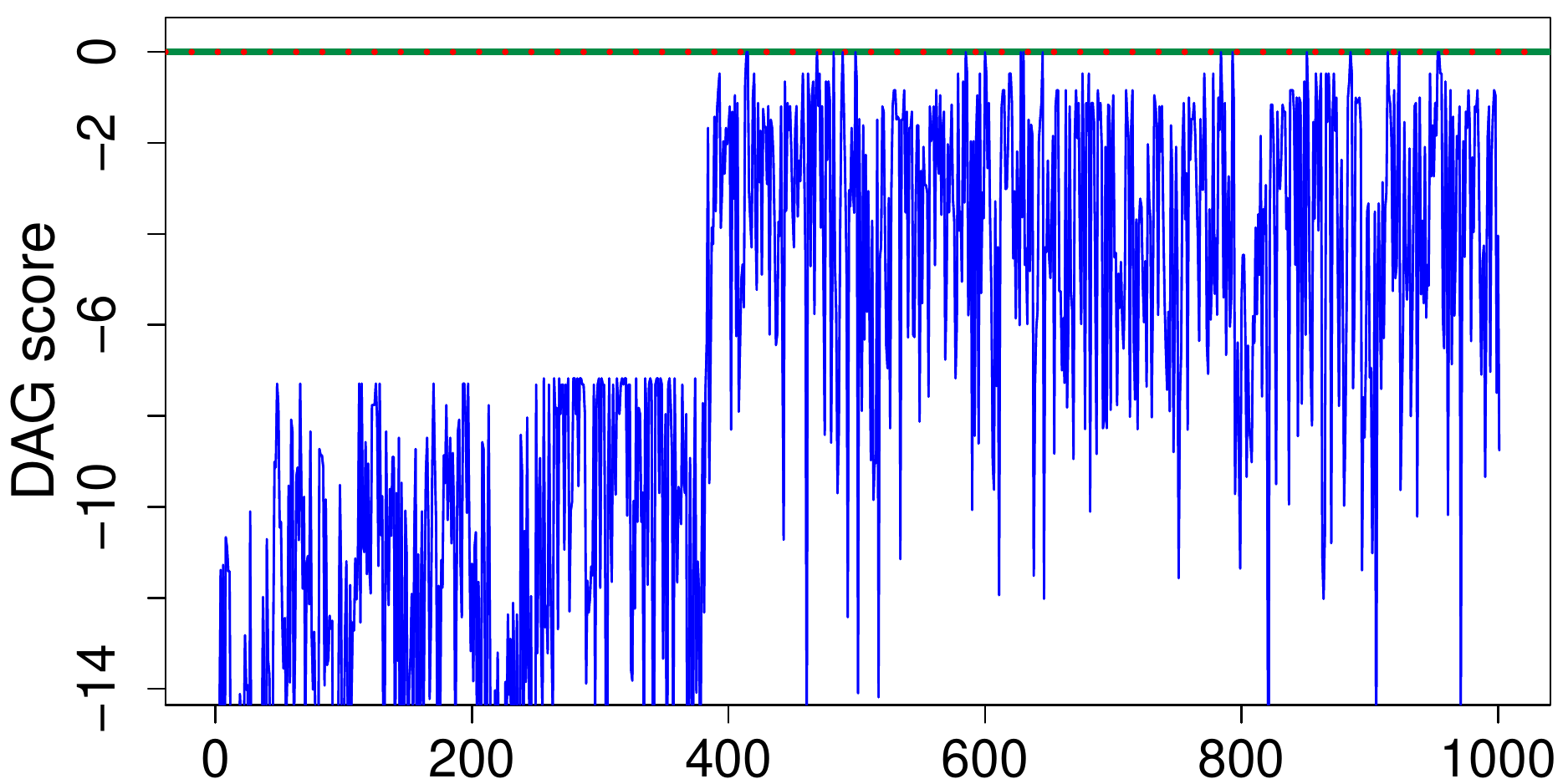} & 
 \includegraphics[width=0.45\textwidth]{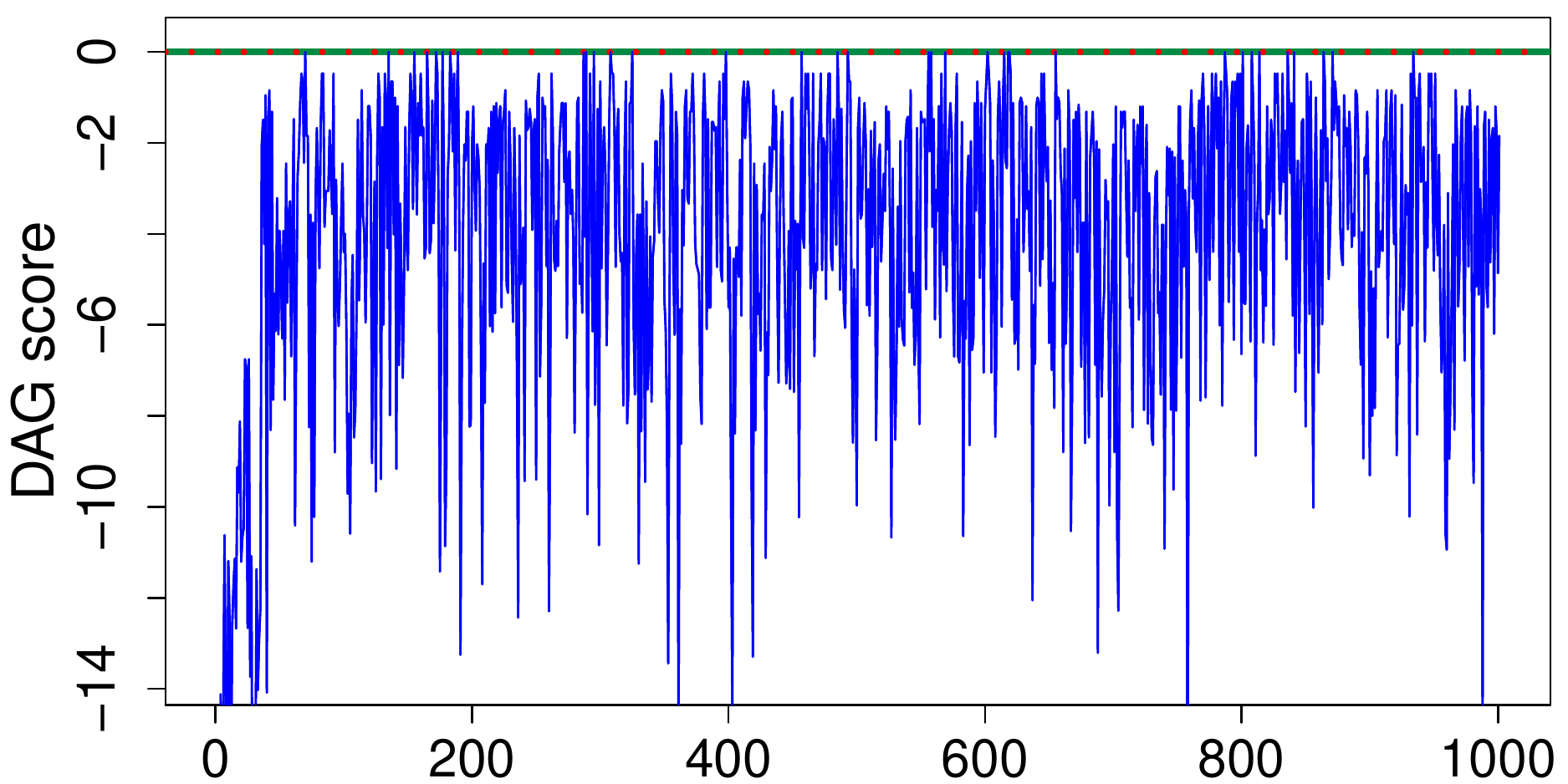} \\
\includegraphics[width=0.45\textwidth]{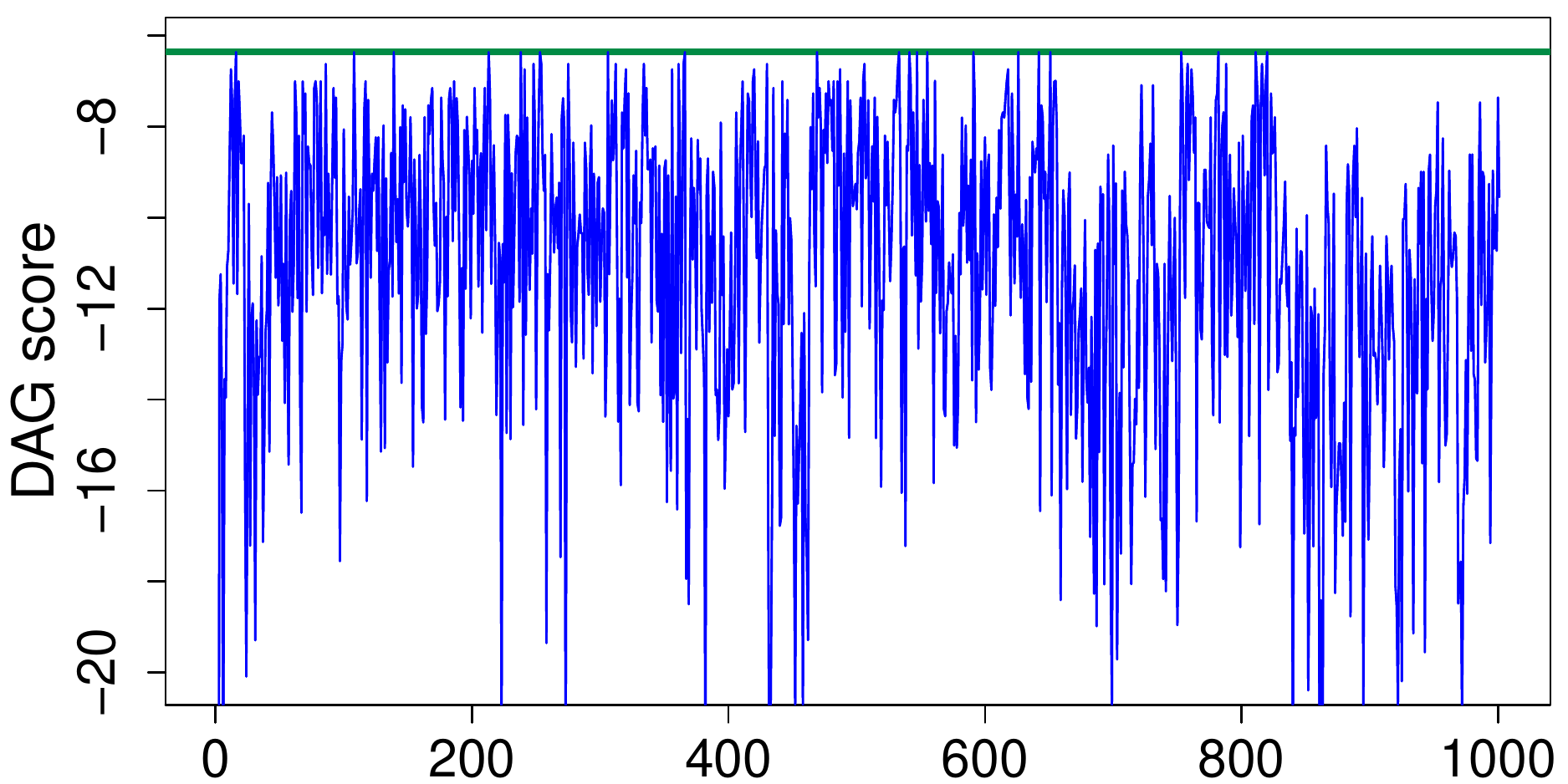} & 
 \includegraphics[width=0.45\textwidth]{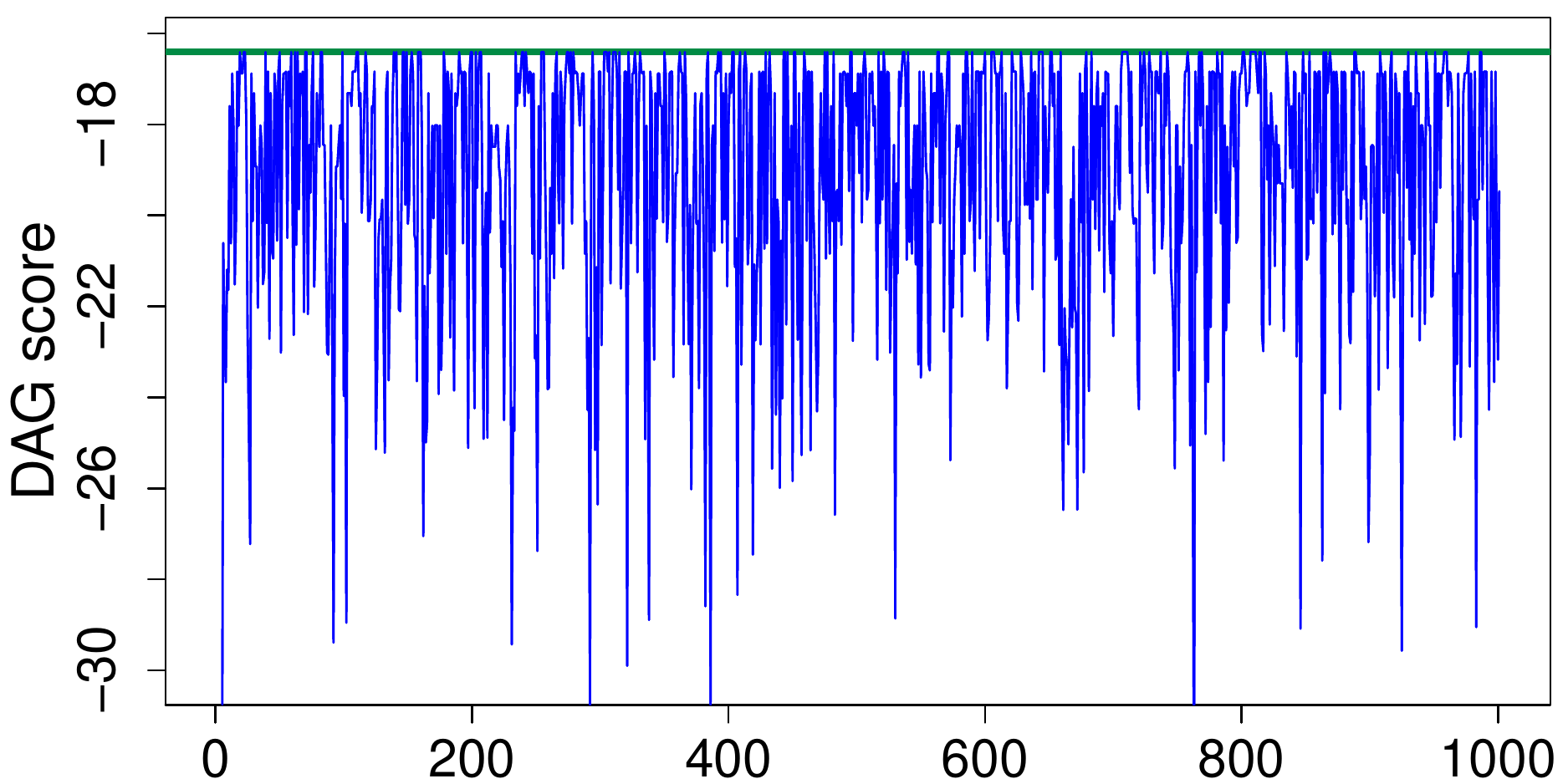}
 \end{array}$
  \caption{A run of half a million steps of the new edge reversal MCMC with different seeds for the Boston Housing data.  The top two reach the region of the global maximum, while the others are some distance away.}
  \label{edgerevBH}
\end{figure}

The difference between pure structure MCMC and its combination with the edge reversal move of \cite{gh08} was not evident for the previous simulated example, since the space is quite small and the chains are already quite long.  Looking at the behaviour of the algorithms on the Boston Housing data instead highlights the improvement.  Now we run the chains for half a million steps and plot the results in \fref{edgerevBH}.  None of the runs are as far away as the worst examples in \fref{structureBH} though one a run still behaves similarly to the other examples of \fref{structureBH}.  Two of the examples in \fref{edgerevBH} are around the global maximum however.  When comparing to partition MCMC, local exploration seems better from the longer runs, but correspondingly long times are spent around each horizontal level.  The latter is due of course to the structure moves but the edge reversal moves increase the chance of large jumps to new levels (and hence finding the global maximum) though they are still rarely successful.    

\begin{figure}
  \centering
$\begin{array}{cc} \includegraphics[width=0.45\textwidth]{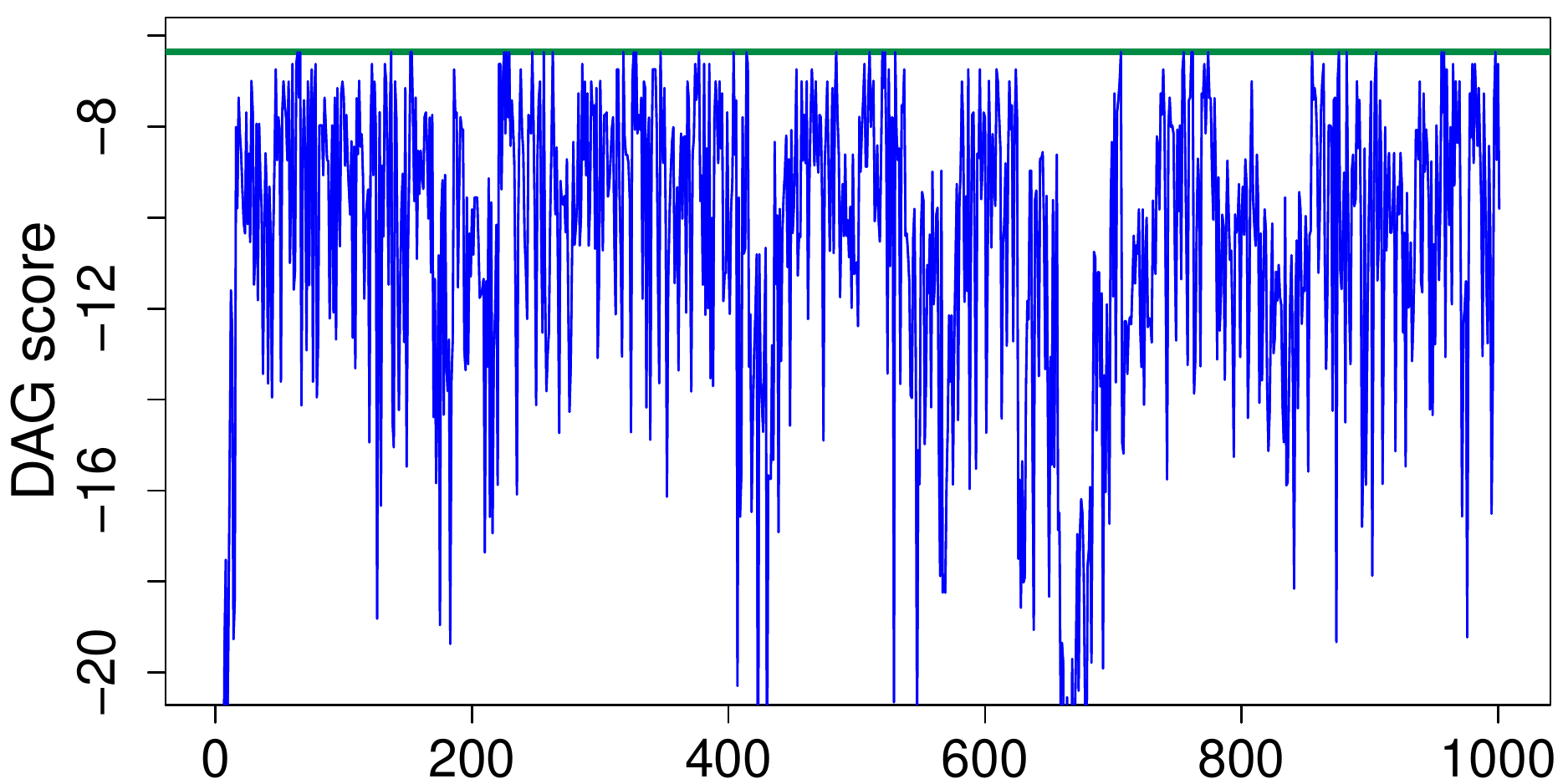} & 
 \includegraphics[width=0.45\textwidth]{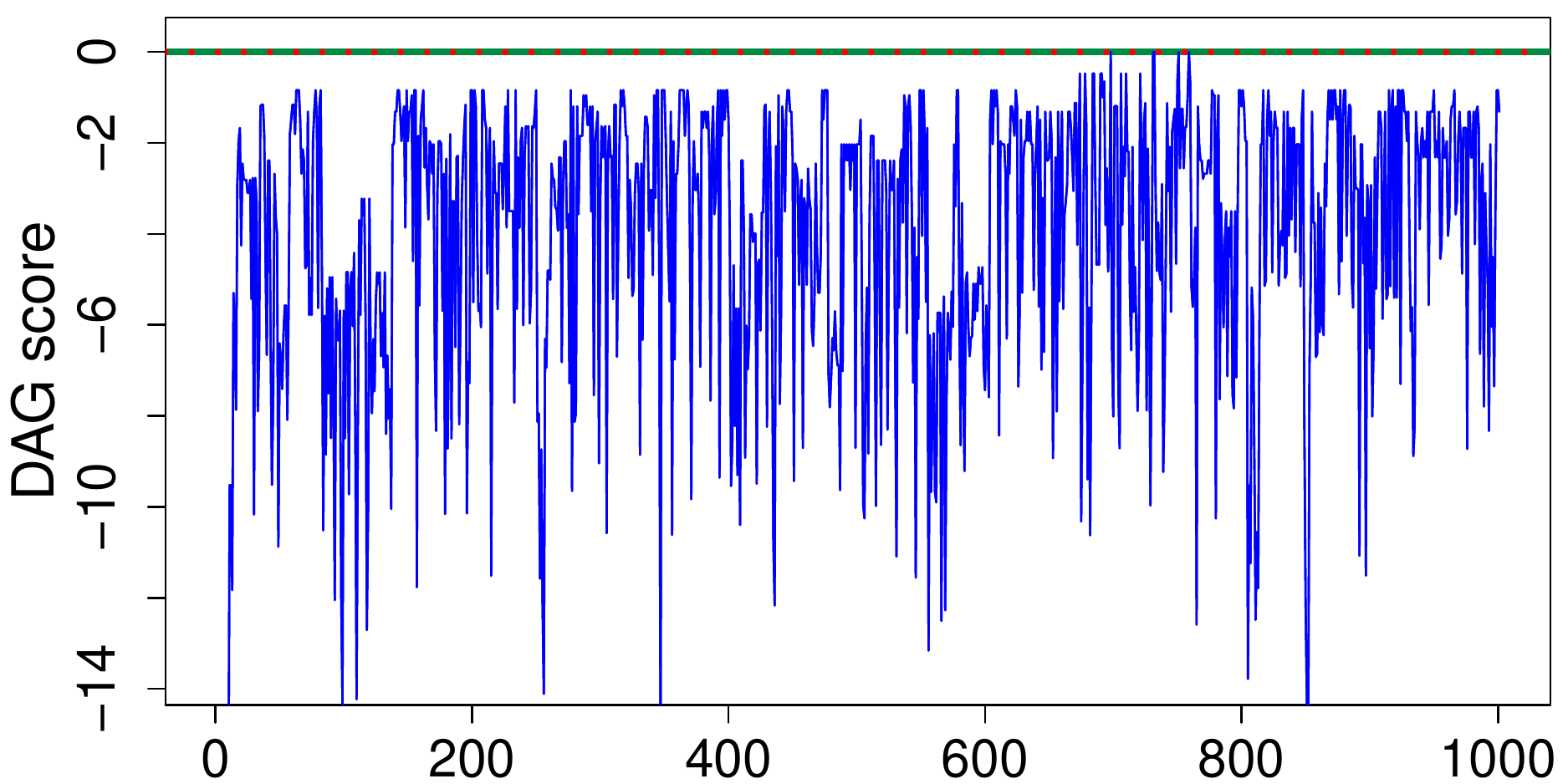} \\
\includegraphics[width=0.45\textwidth]{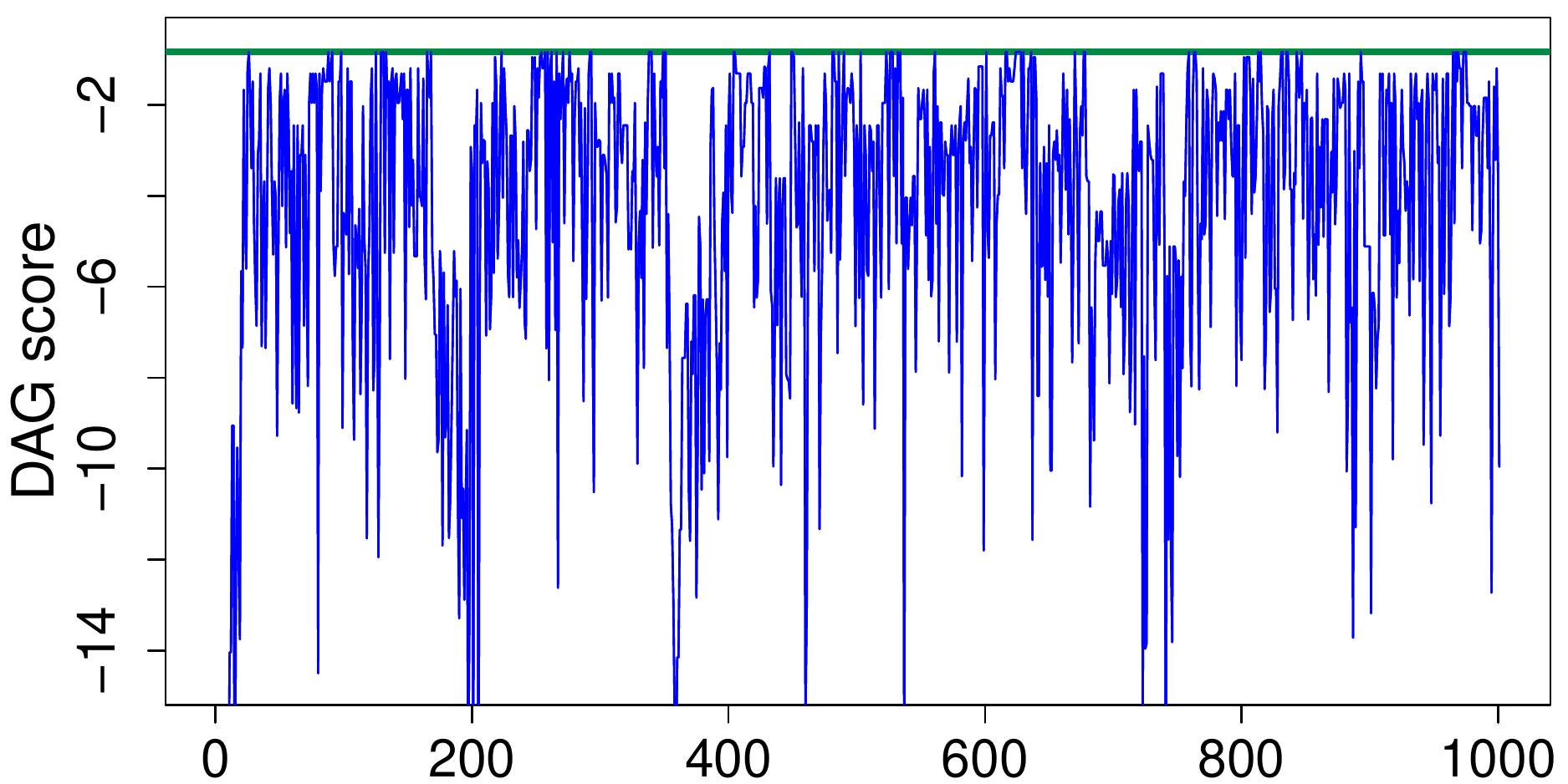} & 
 \includegraphics[width=0.45\textwidth]{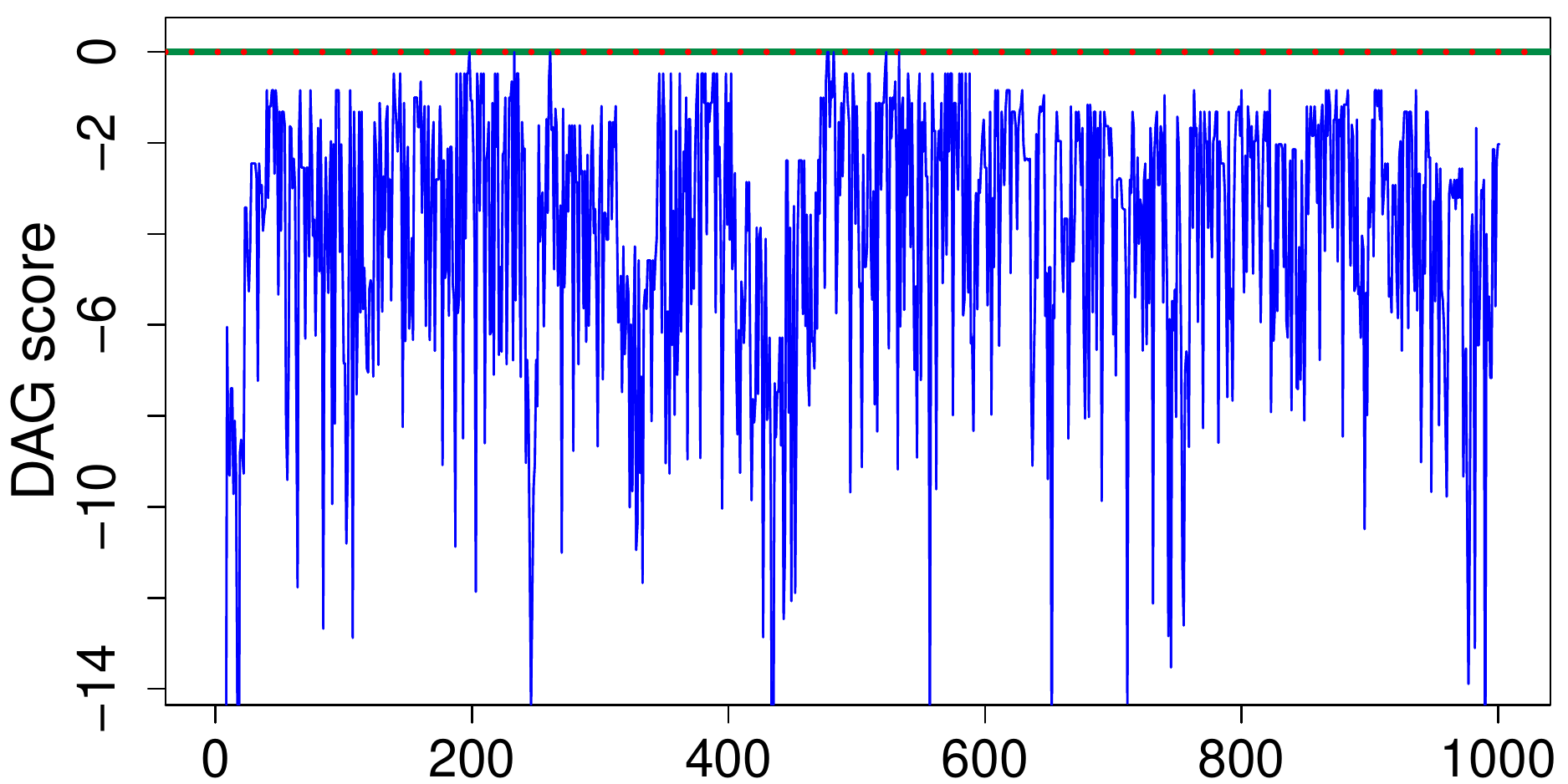}
 \end{array}$
  \caption{A run of 56 thousand steps of partition MCMC with edge reversal with different seeds for the Boston Housing data.  The runs on the right seem to reach the global maximum set at 0, while the first plot remains just a little lower.}
  \label{partitionBHedgerev}
\end{figure}

Combining the edge reversal with partition MCMC instead we run chains of 56 thousand steps as opposed to the 60 thousand before.  Of course the exact timing of each run depends on the acceptance probability and can be quite variable.  Trace plots are shown in \fref{partitionBHedgerev} and they seem to combine the best features of \frefs{partitionBH} and~\ref{edgerevBH}.

\begin{figure}
  \centering
$\begin{array}{cc} \includegraphics[width=0.45\textwidth]{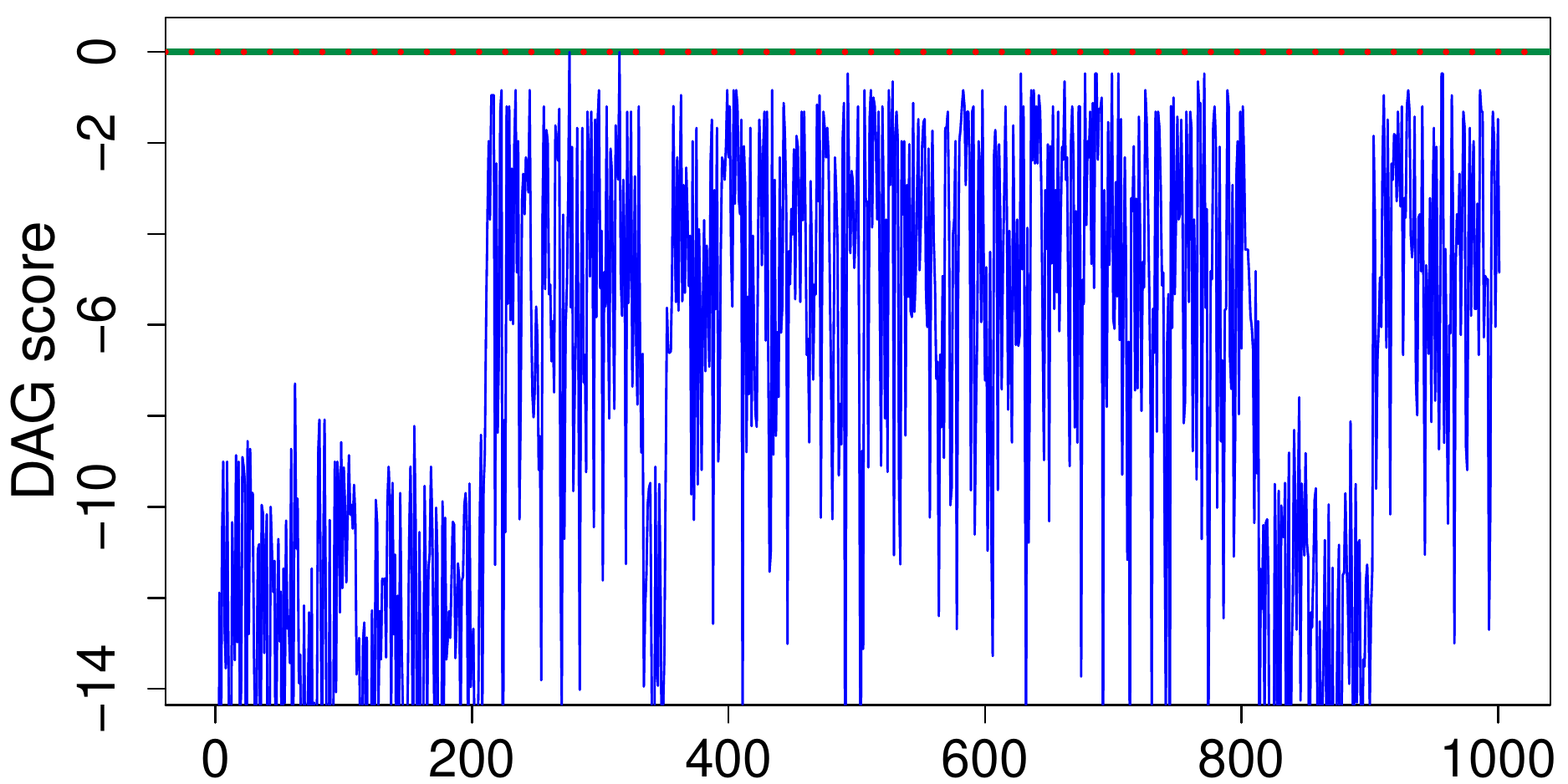} & 
 \includegraphics[width=0.45\textwidth]{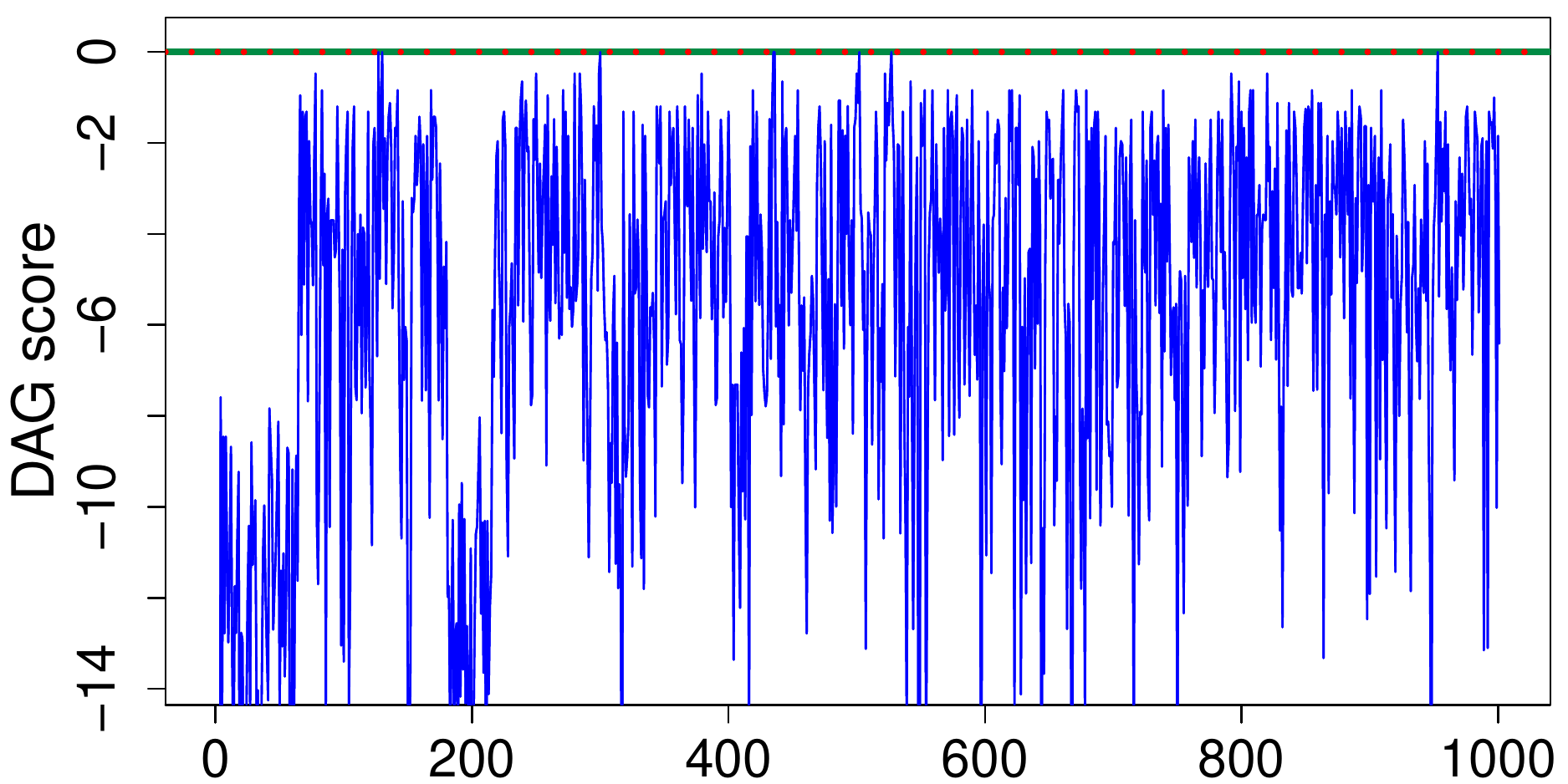} \\
\includegraphics[width=0.45\textwidth]{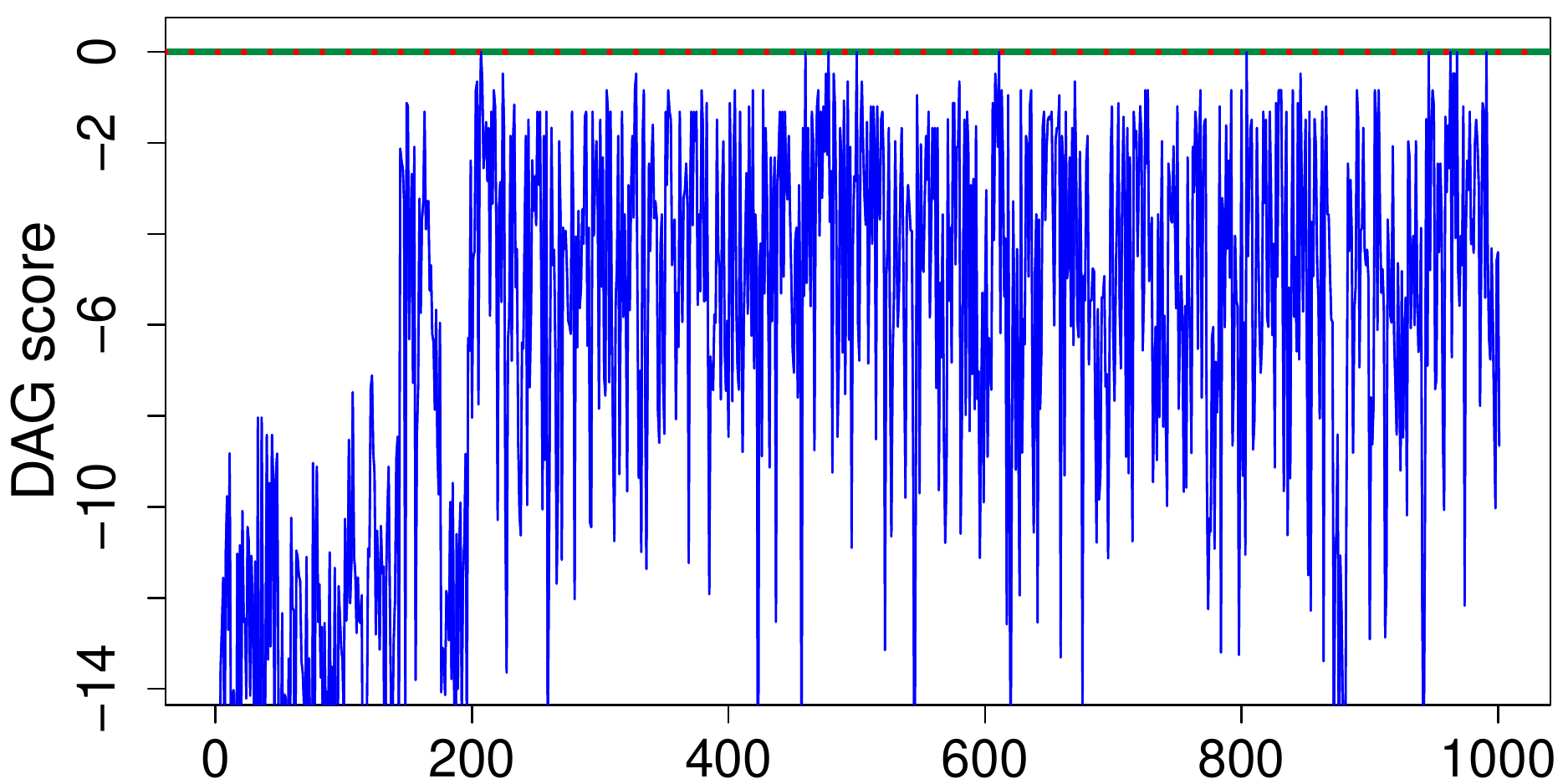} & 
 \includegraphics[width=0.45\textwidth]{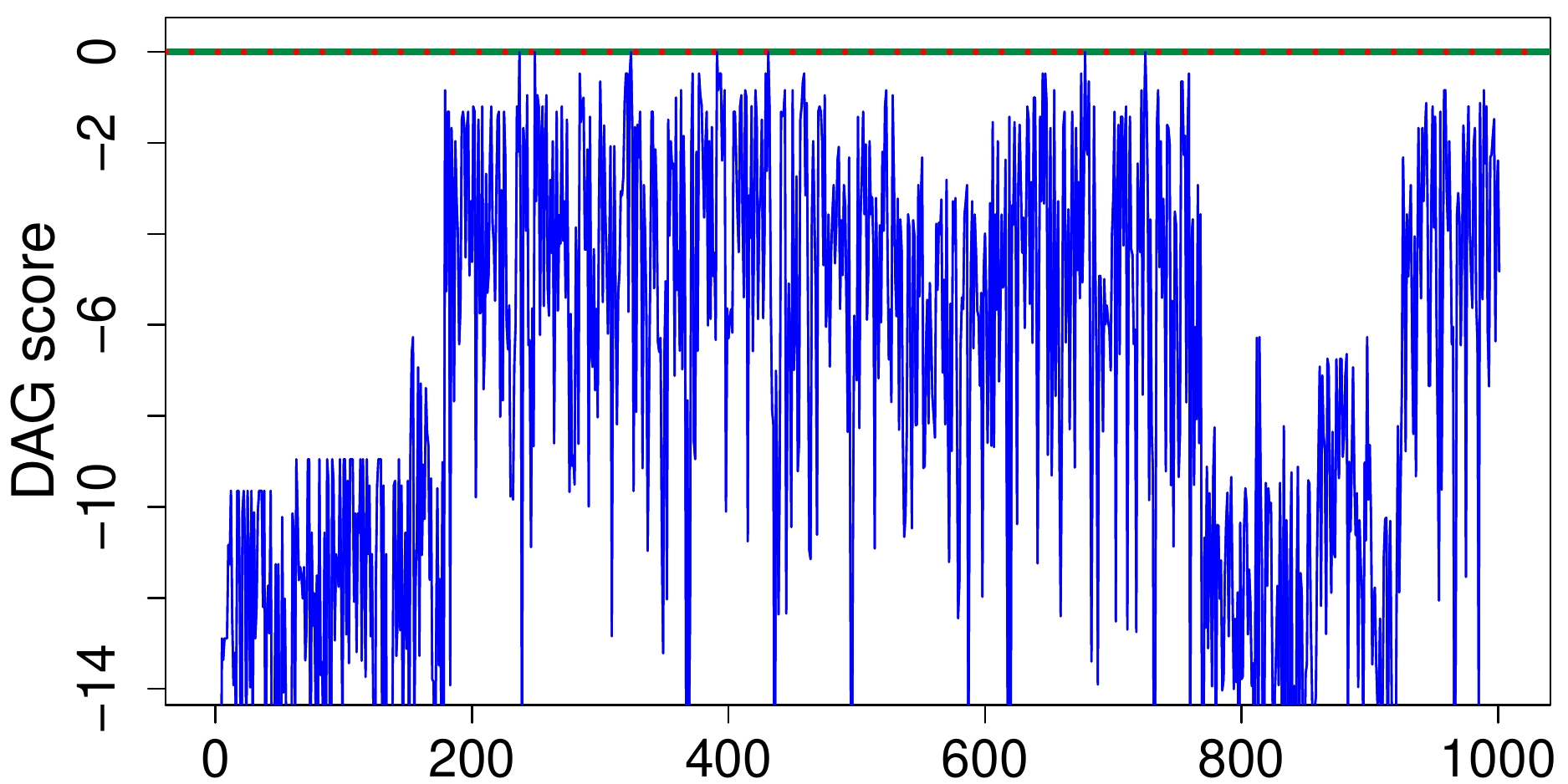}
 \end{array}$
  \caption{A run of 150 thousand steps of order MCMC with different seeds for the Boston Housing data.  All the runs reach the global maximum at 0, but there seems to be a strong propensity to explore a region between -10 and -14.}
  \label{orderBH}
\end{figure}

For comparison, we also run order MCMC for 150 thousand steps, for which trace plots appear in \fref{orderBH}. The performance in finding the maximum is the best, but a relative score region between -10 and -14 seems overly represented, which may be due to the bias.

\begin{figure}
  \centering
\includegraphics[width=0.65\textwidth]{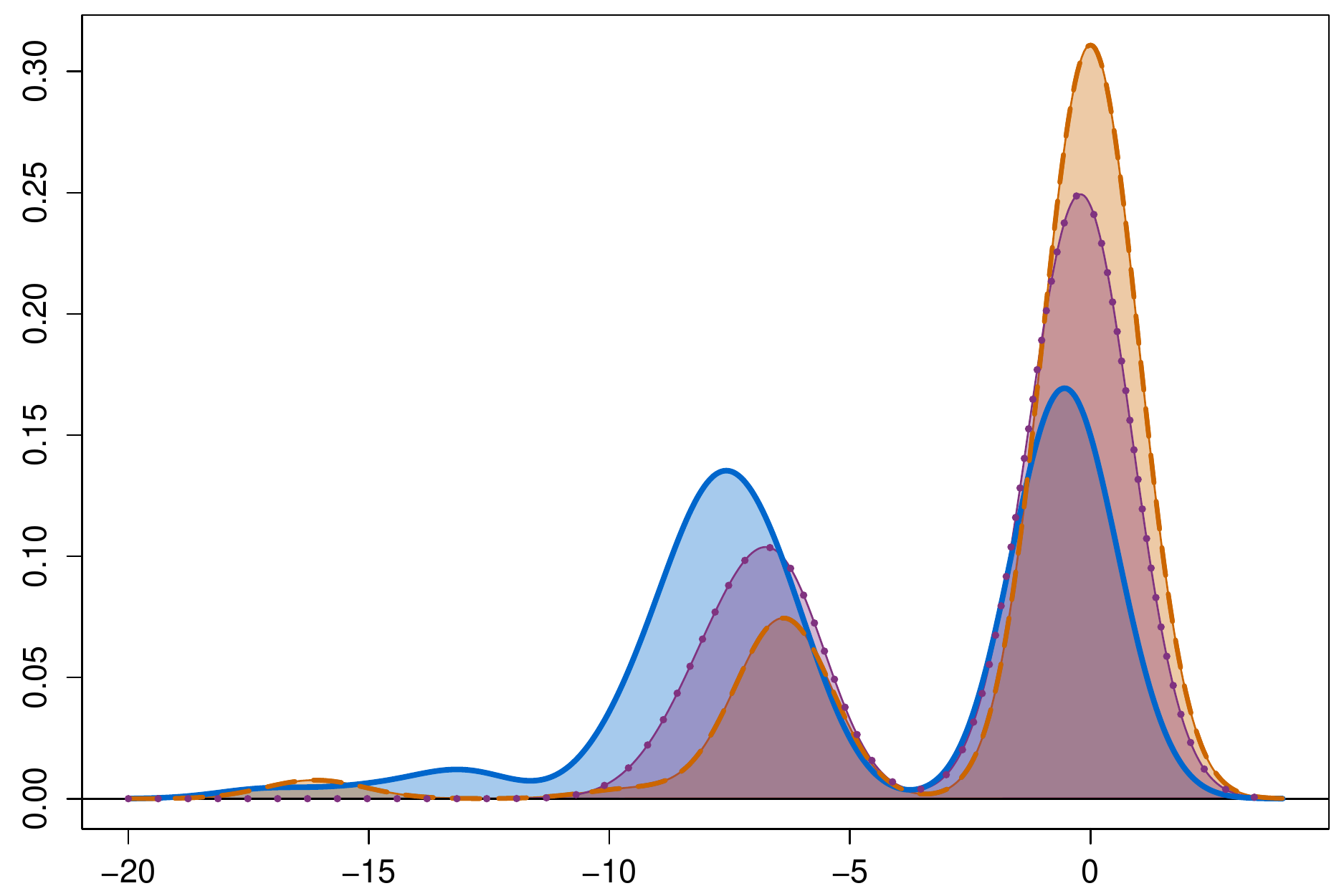} 
  \caption{Density plots of the maximal score found by partition MCMC (solid blue), structure with edge reversal (dashed orange) and partition with edge reversal (dotted purple) for the Boston Housing data.}
  \label{BHdensities}
\end{figure}

To compare the ability of the algorithms to discover the maximum, the four example trace plots presented do not suffice so we run each of the better unbiased methods with 100 different seeds. A density plot of the maximal scores they discover is presented in \fref{BHdensities}.  Edge reversal finds the maximum most often, partition MCMC the least.  Partition MCMC however finds a large range of possible values as opposed to the handful found by edge reversal suggesting that the edge reversal chains follow more similar paths through the score landscape.  Combining partition MCMC with edge reversal provides intermediate behaviour.

\subsection{Simulation from a more connected DAG}

\begin{figure}
  \centering
\includegraphics[width=0.65\textwidth]{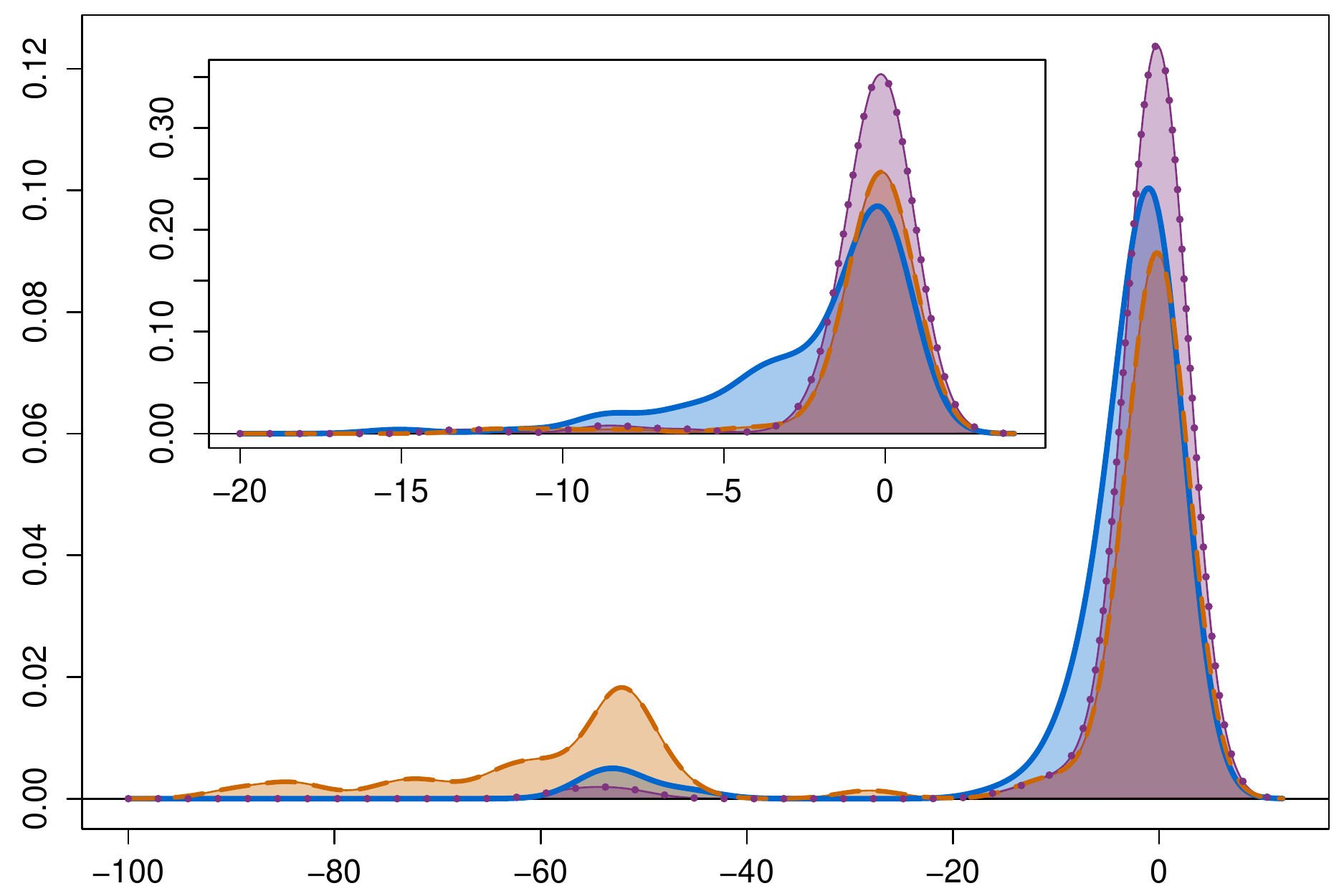} 
  \caption{Density plots of the maximal score found by partition MCMC (solid blue), structure with edge reversal (dashed orange) and partition with edge reversal (dotted purple) for simulated data on 14 nodes.  In the inset, we zoom in the region around 0 with a narrower convolving function.}
  \label{14nodesdensities}
\end{figure}

Keeping $n=14$ we move to a simulation of a more connected DAG with $K=6$.  We sample uniformly a lower triangular $(0,1)$ matrix and remove elements at random from any column with more than $K$ non-zero entries until only $K$ remain.  We then pick a random permutation of the nodes and use the resulting DAG to generate $N=500$ observations following a normal distribution with regression on the parents.  So far the setup is like the Boston Housing data, but with a more connected underlying DAG.  Again we run the different methods with 100 seeds and keep track of the maximal scores discovered by the chains.  The density plot of the maximal scores is an imperfect measure of how good each method is, but indicative.  Keeping half a million steps for structure with the new edge reversal move of \cite{gh08}, instead of timing the structure steps we assume they take negligible time and run partition MCMC for $0.07$ times the number of steps, or 35 thousand.  The combined partition with edge reversal is run for 32 thousand steps each time.  The resulting density of maxima is plotted in \fref{14nodesdensities}.  Despite the slight chain length advantage for edge reversal, it performs worse than partition MCMC with a large tail far away from the global maximum but with a smaller spread around the maximum itself.  The behaviour is a reflection of the relatively small number of different score regions edge reversal discovers compared to the wider spread of partition MCMC.  The clear favourite in \fref{14nodesdensities} is the combination of partition MCMC with edge reversal.  These results suggest that the combination would be the preferred algorithm for inference on DAGs as the size and connectivity increases.

\subsection{Larger simulations}

\begin{figure}
  \centering
$\begin{array}{cc} \includegraphics[width=0.5\textwidth]{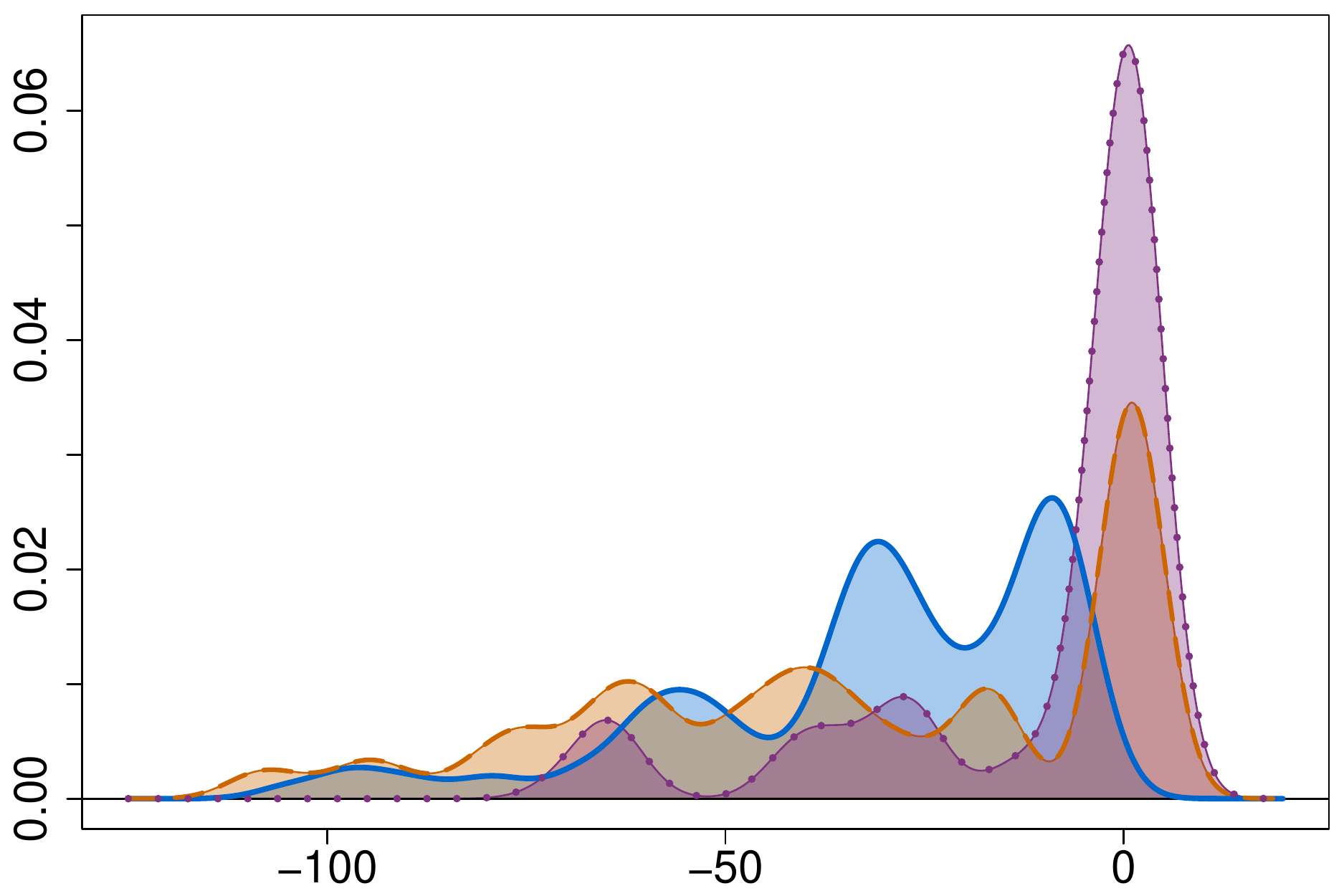} & 
 \includegraphics[width=0.5\textwidth]{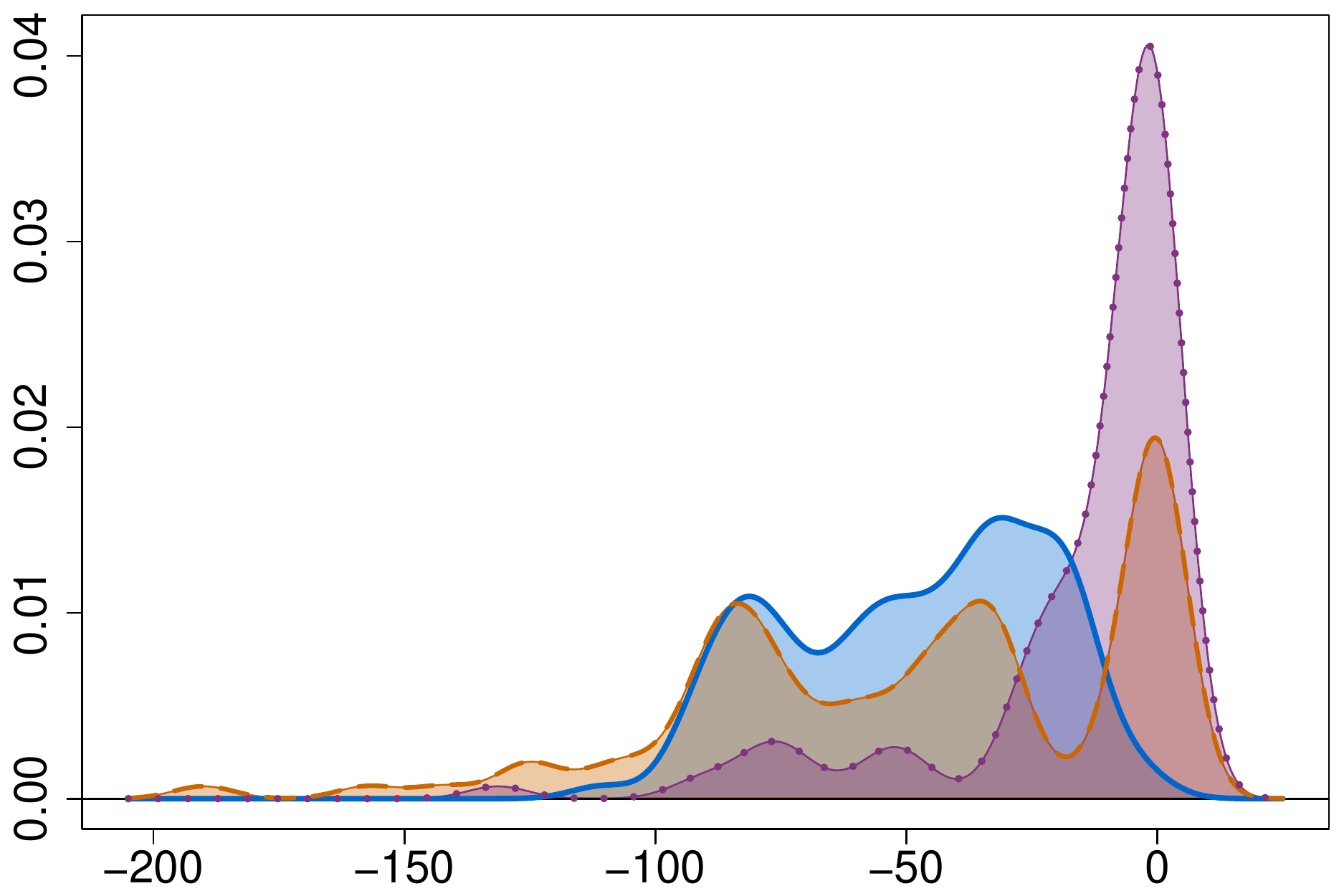} 
 \end{array}$
  \caption{Density plots of the maximal score found by partition MCMC (solid blue), structure with edge reversal (dashed orange) and partition with edge reversal (dotted purple) for simulated data on 18 nodes (left) and 20 nodes (right).}
  \label{1820nodesdensities}
\end{figure}

When moving to larger graphs with $n=18$ and $n=20$ with $N=200$ observations and a limit of $K=5$ on the parents we observe the same improvement by combining edge reversal with partition MCMC as shown in \fref{1820nodesdensities}.

\section{MAP discovery} \label{MAPdiscovery}

\edity{Although we focus on sampling from the posterior, MCMC methods can be adapted to perform a stochastic search for maximum a posteriori (MAP) graphs, or, by replacing the score function appropriately, for (penalized) maximum likelihood discovery. 

A common approach for structure search is greedy hill-climbing, but it has the drawback of stopping in the first local maximum, where MCMC schemes may get only temporarily trapped, as for example in the plateaux visible for the $n=14$ Boston Housing example in \fref{structureBH}.

The complexity of a structure based greedy hill-climbing approach involves testing $O(n^2)$ neighbours at each step to find and move to the best one.  Each neighbour must be scored, which with a fixed limit $K$ on the number of parents is $O(1)$.  After each step, the neighbourhood can be updated in $O(n^2)$ using the ideas of \cite{gc03}.  The structure moves take $O(n^2)$ steps to move through the DAG space leading to a complexity of $O(n^4)$ to find each local optimum.  The overall complexity may be higher if the number of restarts required also grows with $n$.  

A stochastic search based on structure MCMC, using for example simulated annealing, has a complexity of approximately $O(n^5 \ln n)$ \citep{km13} which may grow further if the peakiness of the score landscape likewise grows with $n$.  For practical implementations the coefficients of the complexities play a large role.  Based on the order of complexity though, hill-climbing appears to have the edge; even more so with the improvements in \cite{tba06} in the context of hybrid methods.

For moderate sized problems, as with 14 nodes in \fref{structureBH}, structure MCMC gets trapped in low-scoring local maxima, suggesting that greedy searches would also suffer for larger graphs.  Instead one can search directly in the order or partition space.  The bias due to working in the space of orders rather than the DAG space is not of great concern for MAP learning.  Therefore we start with the simpler permutation space of node orderings $\prec$, of size $n!$.  Each order gets assigned the maximal score of all the DAGs consistent with that node ordering
\begin{equation}
Q(\prec \vert D) = \max_{G\in \prec} P(G\vert D)^{\gamma} \, .
\end{equation}
Instead of hill-climbing through the orders, one can perform a stochastic search with a symmetric MCMC through the space of permutations with acceptance probability
\begin{equation}
\rho = \min\left\{1, \frac{Q(\prec' \vert D)^{\gamma}}{Q(\prec \vert D)^{\gamma}} \right\} \, .
\end{equation}
Throughout the chain the algorithm can keep track of the maximal ordering and hence the maximal DAG discovered.  The power $\gamma$ flattens or sharpens the score landscape and can be tuned to help find the maximum as quickly as possible.  Increasing $\gamma$ as the chain is run corresponds to simulated annealing.

From a complexity perspective, if each move swaps two nodes at a time, it takes $n$ steps for the chain over the space of permutations to become irreducible.  On this irreducible scale of $n$ steps, we assume that the exponential convergence of the MCMC has a rate which is asymptotically independent of $n$.  The chain needs to converge at least to the scale of the inverse size of the space.  To get to $\sim\frac{1}{n!}$ suggests that at least $n\ln n$ irreducible rounds or $O(n^2\ln n)$ MCMC steps are required for good convergence and maximum discovery properties.  The complexity of each MCMC step when carefully weighted as in \aref{app} is $O(n^K)$ leading to an overall behaviour of at least $O(n^{K+2} \ln n)$.  Order and partition MCMC also have the same complexity.  Of course the coefficients may be very different especially due to the possibility of tuning $\gamma$ to speed up the MAP discovery.

For a greedy hill-climbing order search, at each step $O(n^2)$ neighbouring orders are examined.  Moving through this neighbourhood efficiently by only swapping adjacent elements at each step, the cost of scoring each neighbour is still $O(n^K)$.  Moving through the order space requires $O(n)$ steps leading to a minimum complexity of $O(n^{K+3})$ with possible increases depending on how the number of restarts relates to $n$.

The stochastic search may have lower complexity than greedy hill-climbing on orders, and can cope with an uneven score landscape, but its main advantage is the possibility of providing an indication of the confidence that the maximum is the global one.  

Imagine the search uncovers a maximally scoring DAG which belongs to a single order.  To test whether this local maximum may be the global one, a stochastic search is run $Z$ times, discovering the candidate global maximum on $z$ of those runs.  The probability of discovering the maximum on each run would be estimated as $p^{\star}=\frac{z}{Z}$.  Since the search, once converged, is sampling proportionally to $Q^{\gamma}$, it is more likely to hit higher scoring graphs, if they exist.   The probability of doing so on each run should be greater than $p^{\star}$. The probability of missing any higher scoring graphs on any of the runs should be less than about $(1-p^{\star})^Z$.  By modifing $\gamma$, $Z$ and the lengths of the runs, we can reduce the bound to any acceptably low value.

If the candidate maximum happens to belong to $W$ orders the same reasoning leads to a weaker bound of $\left(1-\frac{z}{WZ}\right)^Z$.  Alternatively, one can can turn to the partition space with a unique representation of each DAG, though there may still be equivalent DAGs in other partitions.}

\vskip 0.2in
\bibliography{mcmcdags}

\end{document}